\documentclass[a4paper,fleqn]{cas-sc}

\usepackage[numbers]{natbib}
\usepackage{subfigure}
\usepackage{subfloat}
\usepackage{url}
\usepackage{hyperref}
\usepackage{soul}

\usepackage{xcolor}
\usepackage{enumerate}
\usepackage[normalem]{ulem}

\soulregister\cite7
\soulregister\citep7
\soulregister\citet7
\soulregister\ref7

\def\tsc#1{\csdef{#1}{\textsc{\lowercase{#1}}\xspace}}
\tsc{WGM}
\tsc{QE}
\tsc{EP}
\tsc{PMS}
\tsc{BEC}
\tsc{DE}

\begin{document}
\let\WriteBookmarks\relax
\def\floatpagepagefraction{1}
\def\textpagefraction{.001}
\shorttitle{Dual Reference Age Synthesis}
\shortauthors{Yuan Zhou et~al.}

\title [mode = title]{Dual Reference Age Synthesis}                      
\tnotemark[1]

\tnotetext[1]{This work was supported by the Jiangsu Overseas Visiting Scholar Program for University Prominent Young and Middle aged Teachers and Presidents.}

\author[1]{Yuan Zhou}[type=editor,
                      auid=000,bioid=1,
                      orcid=0000-0002-8224-6068]
\cormark[1]
\ead{zhouyuan@nuist.edu.cn}
\credit{Conceptualization of this study, Methodology, Writing - Original draft preparation}
\address[1]{School of Artificial Intelligence, Nanjing University of Information Science and Technology, Nanjing, China}

\author[2]{Bingzhang Hu}
\ead{bingzhang.hu@ncl.ac.uk}
\credit{Conceptualization of this study, Methodology}
\address[2]{School of Computing, Newcastle University, UK}

\author[1]{Jun He}
\ead{jhe@nuist.edu.cn}
\credit{Writing - Original draft preparation}

\author[2]{Yu Guan}
\ead{yu.guan@ncl.ac.uk}
\credit{Resources}

\author[3]{Ling Shao}
\cormark[2]
\ead{ling.shao@ieee.org}
\address[3]{Inception Institute of Artificial Intelligence (IIAI) Abu Dhabi}
\credit{Supervision, Writing - Review and Editing}

\cortext[cor1]{Corresponding author}
\cortext[cor2]{Principal corresponding author}

\begin{abstract}
Age synthesis methods typically take a single image as input and use a specific number to control the age of the generated image. In this paper, we propose a novel framework taking two images as inputs, named dual-reference age synthesis (DRAS), which approaches the task differently; instead of using ``hard'' age information, \textit{i.e.} a fixed number, our model determines the target age in a ``soft'' way, by employing a second reference image. Specifically, the proposed framework consists of an identity agent, an age agent and a generative adversarial network. It takes two images as input - an identity reference and an age reference - and outputs a new image that shares corresponding features with each. Experimental results on two benchmark datasets (UTKFace and CACD) demonstrate the appealing performance and flexibility of the proposed framework.
\end{abstract}



\begin{keywords}
age synthesis \sep dual reference \sep ``soft'' age information \sep conditional generative adversarial network
\end{keywords}

\maketitle

\section{Introduction}

Age synthesis, also known as face aging and rejuvenation or age progression and regression, aims to predict aging or rejuvenating effects on an individual's facial images, while preserving personality features. It has received significant research interest in recent years due to its importance for a wide range applications, \textit{i.e.} finding missing people, face verification, security surveillance, entertainment, \textit{etc.} Before the emergence of generative adversarial network (GAN) \cite{goodfellow2014generative}, popular age synthesis algorithms focuse on the shape or texture analysis, which is related to craniofacial growth or skin aging in age progression \cite{ricanek2006craniofacial}, or consider shape and texture synthesis simultaneously \cite{fu2010age}. With the breakthrough of GAN, synthesis methods based on GAN have yielded great progress. Conventional age synthesis methods are categorized into three groups: methods based on physical model, prototype based methods and GAN based methods. Physical model based methods use a parametric anatomical model to describe the face aging procedure, including how the facial skin changes, the physical mechanism on facial cranial growth, and facial muscle changes\cite{mark1980wrinkling,o1997three,o19993d,suo2009compositional,tazoe2012facial}. However, these physical model based methods are computationally expensive and complex \cite{wang2016recurrent,zhang2017age,wang2018face}. In prototype-based methods, prototypes are learned to define the salient feature at different ages, and then the age transformation is depicted as the discrepancy between two prototypes \cite{tiddeman2001prototyping,benson1993extracting,kemelmacher2014illumination}. This learned age transformation is then applied to an input face to produce the corresponding aging effects. However, the prototypes are simply averages of facial features and are thus unable to preserve identity information \cite{liu2004image}. The GAN-based methods typically combine a GAN with an encoder for age synthesis, where the GAN is used to synthesize an image with the identity feature learned by the encoder  \cite{yang2017learning,antipov2017face}. The age of synthesized image can be controlled by transforming a fixed target age to a one-hot vector. For optimal performance, GAN-based methods require a huge volume of pair-wise images (\textit{i.e.} facial images of the same person across a large age span) which are difficult and often infeasible to obtain. 

\begin{figure*}
    \centering
    \subfigure[]{\includegraphics[width=0.45\textwidth]{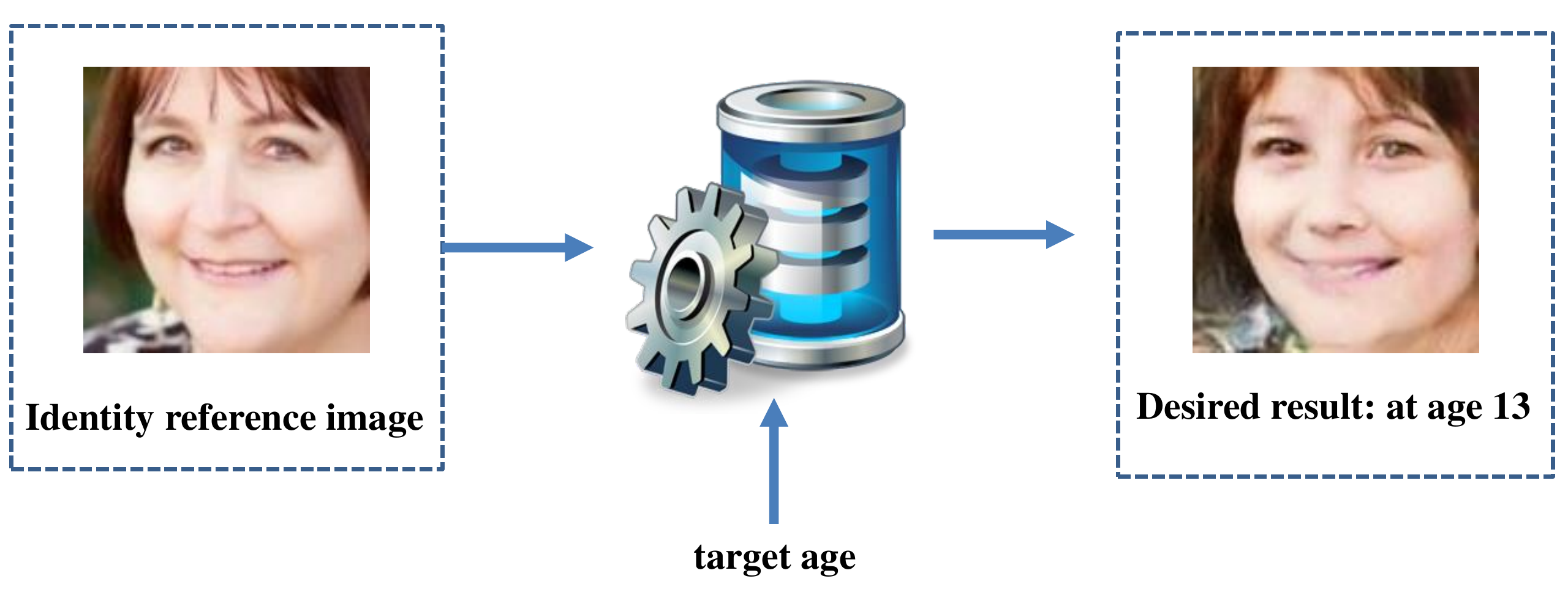}}
    \quad
    \subfigure[]{\includegraphics[width=0.45\textwidth]{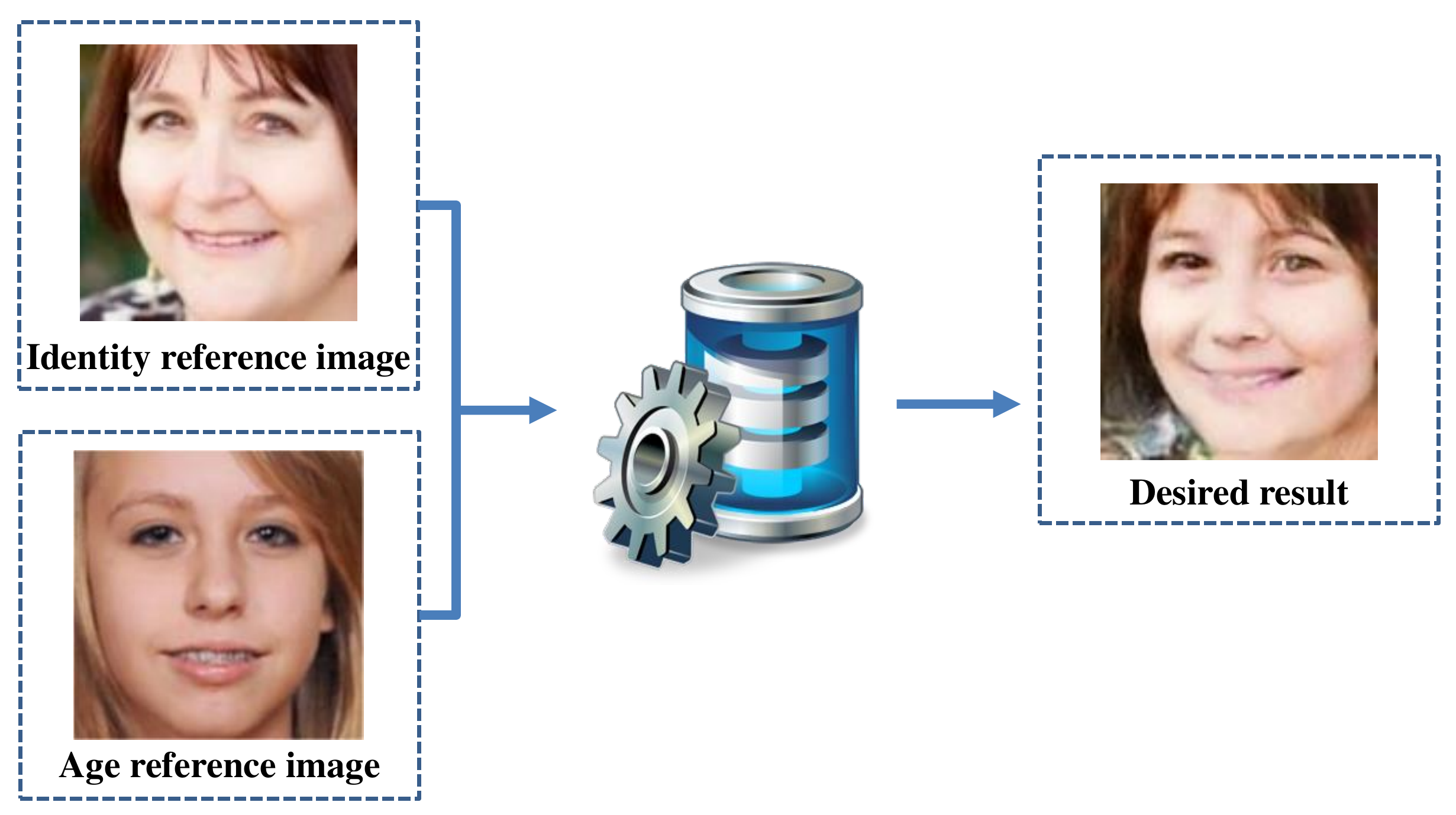}}
    \caption{Comparison between a conventional age synthesis framework and our proposed dual-reference age synthesis framework. (a) Conventional GAN-based age synthesis framework: a one-hot-vector as an age feature is transformed from the target age, then an identity reference image and the one-hot-vector are fed into an generator to synthesize a facial image. (b) Dual-reference Age Synthesis framework: the framework synthesizes facial images in a dual-reference manner, where one image is the identity reference image and the other is the age reference image.}
    \label{fig:comparision}
\end{figure*}

Existing works use either a fixed number or descriptions like ``young'' and ``old'' to represent the age information desired in the output. However, different people may look different ages, even if they're not, while different observers will have different understandings of ``young'' and ``old''. This naturally rises the question: are current depictions adequate enough for accurately describing human age? Conventional methods use a number to control the age of a synthesised image, as shown in Figure \ref{fig:comparision}(a). However, one drawback to this is that a single number does not fully capture human perceptions of age. As the old saying goes, ``a picture is worth a thousand words'', a facial image provides far more age information than a number or an average depiction does. Thus, we propose a new task: can we use someone else image, at a specific age, to set the target age for face age synthesis? To tackle this task, we propose a novel framework, in which an age reference image, in addition to the identity reference image, is input to reflect the target age. We refer to the proposed framework as dual-reference age synthesis (DRAS), as shown in Figure \ref{fig:comparision}(b).

The contributions of this paper are three-fold:
\begin{enumerate}
    \item \textit{Task:} We propose a new task to synthesize images of one input face at a similar age to that of a second input image. The proposed task successfully addresses the problem of a single number  not being able to effectively represent human age.
    \item \textit{Framework:} A unified framework is proposed to tackle the new task. Using the mechanisms of two independent discriminators, the proposed framework can generate images with the similar age as the age reference image, while preserving identity information.
    \item \textit{Performance:} Extensive experiments and detailed analyses are conducted on two benchmark datasets: UTKFace and CACD. Our model achieves the best performance among compared methods for age synthesis, and is more feasible, especially for tasks lacking ground truth or pair-wise datasets.
\end{enumerate}

\section{Related Work}

Physical models based on facial landmarks have been used for real world age synthesis tasks since as far back as 2002. Lanitis \textit{et al.} investigated three aging formulations and used 50 raw parameters to describe aging effects on facial appearance \cite{lanitis2002toward}. Mukaida and Ando \cite{mukaida2004extraction} extracted and seperated facial wrinkles and spots for age synthesis by analyzing the properties of pixel distributions in local areas. Gandhi \textit{et al.} intorduced a real-world age synthesis system \cite{gandhi2004method} which mainly focused on texture synthesis by considering both signature images and regression-based age prediction. Ramanathan \textit{et al.} defined a craniofacial growth model, including a shape aging model and a texture aging model, to characterize adult facial shape and textural variations occurring with age \cite{ramanathan2006modeling}. Fu and Zheng \cite{fu2006m} presented the M-Face framework that the shape caricaturing is integrated for associated shape deformation. Since these models consider aging effects as a continuous procedure, \textit{i.e.} from ``young'' to ``old'', dense long-term face aging sequences are needed to obtain the ``best'' model. To tackle the lack of sufficient long-term face aging sequences for model learning, Suo \textit{et al.} attempted to model facial muscle patterns from available short-term aging databases by a proposed concatenational graph evolution aging model. Later on, they decomposed human faces into mutually interrelated sub-regions under anatomical guidance, and proposed an aging model by connecting sequential short-term patterns following the Markov property of the aging process \cite{suo2012concatenational}. Recently, facial textures comprising skin texture details around facial meso-structures (\textit{e.g.} eyes, nose and mouth) have been used to represent the aging effect, surpassing prior work \cite{kaur2017photo}.

Prototype model based methods developed almost in parallel to the physical model based methods. Rather than modeling continuous aging effects, like the physical model based methods do, prototype model methods divide age ranges into discrete. Tiddeman \textit{et al.} proposed wavelet-based methods for prototyping facial textures and shapes, and for artificially transforming the age of facial images \cite{tiddeman2001prototyping}. Not focusing on facial shape or texture, a novel technique named image-based surface detail transfer (IBSDT) was proposed. In IBSDT, aging effects in facial image are obtained by transferring the bumps from an old person's skin surface to a young person's face. Though IBSDT is simple to implement, it needs to manually add markers to the boundaries and the feature points \cite{liu2004image}. These early prototype based methods used average facial shape, textures, and bumps to describe aging transformation directly. Then, Kemelmacher-Shlizerman \textit{et al.} proposed an illumination-aware age progression approach(IAAP) to compute average image subspaces, and use these average depictions (shape, texture) to yield an age progressed result \cite{kemelmacher2014illumination}. Thereafter, trying to maintain personality, Shu \textit{et al.} \cite{shu2015personalized} proposed a coupled dictionary learning(CDL) method. In CDL, a dictionary for each age group is learnt and every two neighbouring dictionaries are learnt jointly. However, this method still has ghost artifacts as the reconstruction residual does not evolve over time \cite{wang2016recurrent}. After that, Shu \textit{et al.} proposed a Kinship-Guided Age Progression (KinGAP) approach which can generate personalized aging images by computing average ageed faces taking the senior family members as a prior guidance \cite{shu2016kinship}. Bukar \textit{et al.} proposed a novel algorithm \cite{bukar2017BenNeedham,bukar2017facial} hybriding the active appearance models(AAM) \cite{cootes2001active} and face patches method to produce aing images with fine facial texture details which eliminated illumination differences. First, an invertible model of age synthesis is developed using AAM and sparse partial least squares regression (sPLS). Then the texture details of the face are enhanced using the patch-based synthesis approach.

Physical and prototype model based methods have dominated age synthesis in the last decade, however, their disadvantages, \textit{i.e.} computational cost, complexity and missing facial details, have hindered high quality synthesis. The conditional generative adversarial network (cGAN) \cite{mirza2014conditional} broke this challenge. cGAN introduces condition into the original GAN to control the generated results, making end-to-end age synthesis possible \cite{mirza2014conditional}. Given a uniform noise $z$ and a condition $y$, they are combined in a joint hidden representation and are mapped to data spaces as $\tilde{x}$ by a generator. $y$ is fed into a discriminator as an additional input with a real data $x$ or a fake data $\tilde{x}$. The discriminator tries to tell them apart, while the generator is trained to prevent this. Moreover, Makhzani \textit{et al.} proposed an Adversarial Autoencoder (AAE), which can be used to learn identity features \cite{makhzani2015adversarial}. The controllable character of cGAN and the latent vector learning ability of AAE inspired GAN-based methods, which use AAE to learn identity features and cGAN to generate aged facial image \cite{antipov2017face,lample2017fader,liu2017face}. Different from physical and prototype model based methods, GAN-based methods use a number to represent age. To disentangle personality and age, Zhang \textit{et al.} proposed a conditional adversarial auto-encoder (CAAE), which includes an encoder and a GAN \cite{zhang2017age}. Personal identities were determined by mapping the original face image to a latent vector via the encoder, then these identities and a corresponding numeral (age) were fed into the GAN to synthesize facial images. Antipov \textit{et al.} proposed an Age Conditional Generative Adversarial Network (Age-GAN) which used Facenet to optimize latent identity vectors \cite{antipov2017face}. Age-GAN can be considered a type of CAAE. Recently, focusing on identity preservation, Wang \textit{et al.} proposed an identity-preserving conditional generative adversarial networks (IPCGANs) using an age classifier to force the generated face to be within the target age group \cite{wang2018face}. To obtain more realistic images, Li \textit{et al.} proposed a Wavelet-domain Global and Local Consistent Age Generative Adversarial Network (WaveletGLCA-GAN).  On the other hand, there are GAN-based methods which don't explicitly model face aging synthesis, however, they are feasible  for the task, \textit{e.g.} Expression Generative Adversarial Network (ExprGAN) \cite{ding2017exprgan} and StarGAN \cite{StarGAN2018}. To the best of our knowledge, these methods still use a specific number to describe age group information and require pair-wise or annotated data. It worth noting that our work seems very related to IP-GANs \cite{bao2018towards}, where the latter uses the Gaussian distribution to regularize all attributes' features. However, age doesn't follow a Gaussian distribution, which means IP-GANs is not a suitable choice for age synthesis.

\section{Proposed method}

In this section, we first describe the framework of our proposed method. Two main modules of the framework are discussed in Sec.3.2 and Sec.3.3, respectively. Finally, the objective functions are introduced.  

\subsection{Overview}

\begin{figure}
    \centering
    \includegraphics[width=0.45\textwidth]{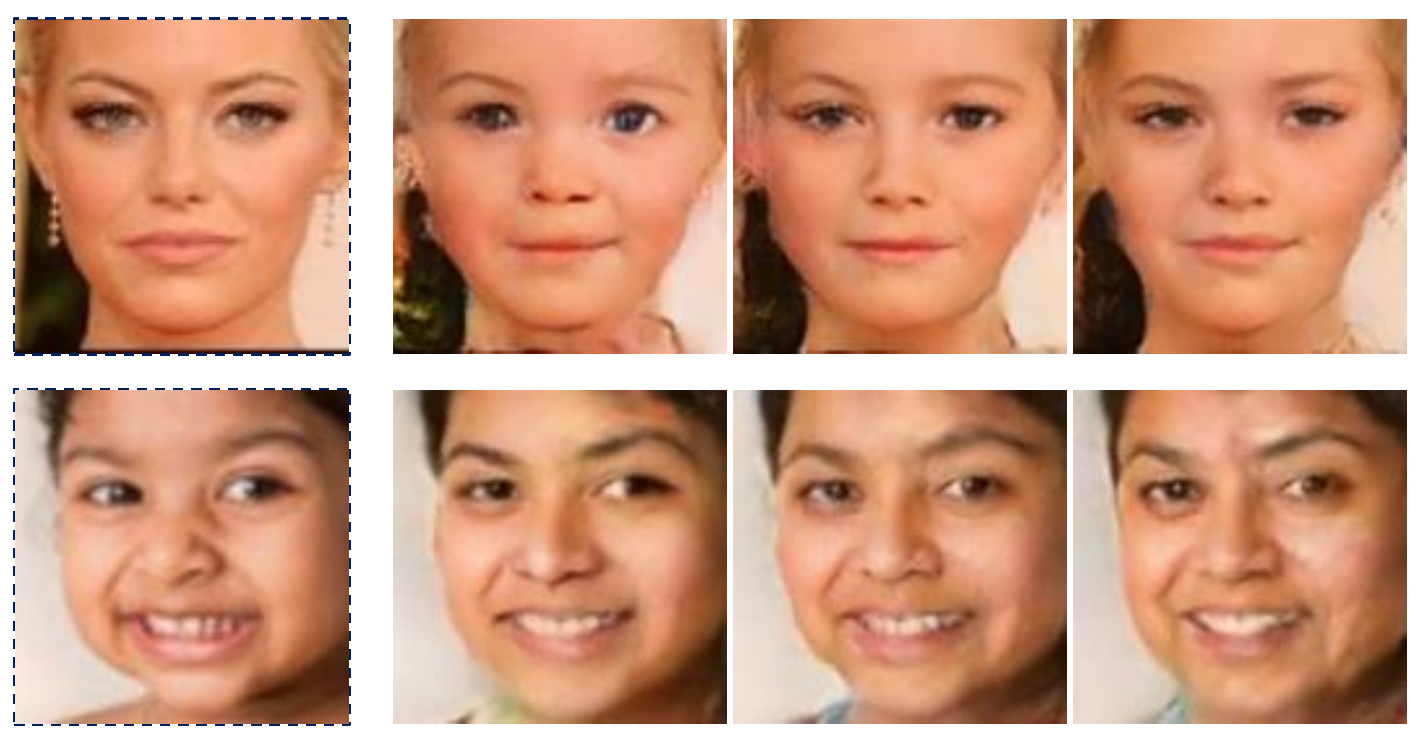}
    \caption{Demonstration of our age synthesis results (images with black dotted boxes are the original ones)}
    \label{fig:intro}
\end{figure}

Given an arbitrary image, can you imagine what will (did) he/she look like in the future (past)? Figure \ref{fig:intro} shows an example of the age progression/regression results. The input images (with black dotted boxes) are manipulated into ``child'', ``young'' and ``old''.

 Different people of the same age often have different age appearances. Therefore, rather than providing a fixed numerical ``hard'' age, it is more reasonable to refer to the ``soft'' version of age information extracted from a facial image via a deep encoder network. Our DRAS framework consists of three parts: an age agent, an identity agent and a GAN. The age and identity features are learned by means of the age and identity agents, respectively. The GAN is used to synthesize photo-realistic facial images. 

\begin{figure*}
    \centering
    \includegraphics[width=0.7\textwidth]{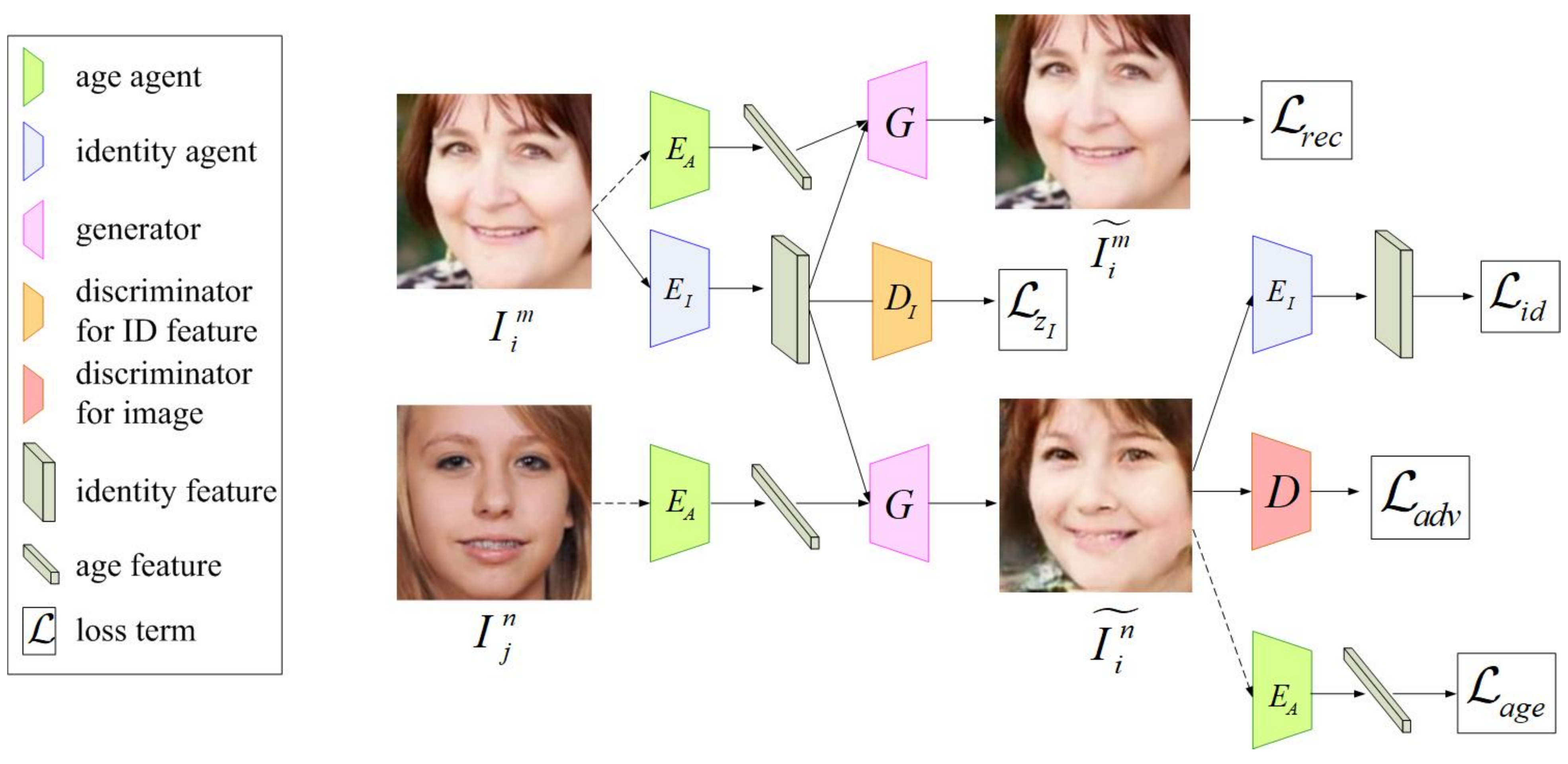}
    \caption{The framework of the proposed age synthesis method. The two raw reference images are taken as input to synthesize images which share the same identity information with the identity reference images and having the same age information with the age reference images.}
    \label{fig:framework} 
\end{figure*}

Figure \ref{fig:framework} describes the framework of our proposed method. For convenience, we define $I_{i}^{m}$ as the identity reference image of the individual with identity $i$ at age $m$, and $I_{j}^{n}$ as the age reference image of the individual with identity $j$ at age $n$. We assume that the face image is sampled from two low dimensional manifolds: the age manifold and identity manifold, where the identity and age change smoothly along their respective dimensions. The two raw reference images are first projected onto the identity and age manifolds, respectively, via the identity agent $E_I$ and the age agent $E_a$. Subsequently, the identity and age features are sampled from these two manifolds, respectively. Moreover, a discriminator $D_I$ is coupled with the identity agent to ensure that the identity features follow a uniform distribution. Then, the identity and age features form a joint feature, which is fed into the generator. Finally, the generator synthesizes a facial image which not only shares the same identity feature as the identity reference image but also shares the same age feature as the age reference image. The hybrid loss function is used to optimize our model. Five losses are included in the hybrid loss function: a reconstruction loss $\mathcal{L}_{rec}$, two adversarial losses $\mathcal{L}_{Z_I}$ and $\mathcal{L}_{adv}$, and two preservation functions $\mathcal{L}_{id}$ and $\mathcal{L}_{age}$. For detailed information, please refer to the following discussions.

\subsection {Age Agent}

An age agent is designed for the proposed framework based on the deep expectation of apparent age (DEX) \cite{Rothe-ICCVW-2015,Rothe-IJCV-2016}, pretrained on ImageNet \cite{deng2009imagenet}. Two fully-connected layers are introduced, while removing the last fully-connected layer from the original DEX. The sizes of these two new fully-connected layers are 1024 and 50, and the 50-dimensional output of the age agent is the final age feature. Furthermore, an image with size $224\times224$ is required as the input age reference image.

{\bf Age preservation:}
The image generated by our DRAS should has the similar age as the reference image, which means the age feature difference between the two should be as small as possible. Here, we use the age preservation loss to describe the similarity of the age feature, referring to Equation~\eqref{eq:AGE_loss}. 
\begin{align} \label{eq:AGE_loss}
\mathcal{L}_{age}=||E_A(I_{j}^{n})-E_A(\tilde{I_{i}^{n}})||_2,
\end{align}
where $E_{A}(\cdot)$ is the age feature, and $\tilde{I}_{i}^{n}$ is the synthesized image. 

In contrast to conventional methods, by introducing the age preservation loss, the age agent is trained in an unsupervised manner without age annotation but only with the ground-truth age feature $E_A(I_{j}^{n})$. When training our DRAS, parameters of the last two full-connected layers are optimized to better learn age features through back-propagation of the age preservation loss. Moreover, using the 50-dimensional feature rather than the conception features (congregated multi-layer outputs of deep networks) makes our framework light-weight  \cite{yang2017learning,ding2017exprgan}. 

\subsection{Identity Agent}

The identity agent consists of an encoder $E_I$ and a discriminator $D_I$, whose architectures are adapted from \cite{zhang2017age}. The encoder takes a $128\times128\times3$ image as input. 

{\bf Reconstruction:}
In order to extract identity features from identity reference images without pair-wise or labeled training data, a reconstruction loss is used:
\begin{align} \label{eq:rec_loss}
\mathcal{L}_{rec}=||I_{i}^{m}-\tilde{I}_{i}^{m}||_1,
\end{align}
where $\tilde{I}_{i}^{m}=G(E_I(I_{i}^{m}),E_A(I_{i}^{m}))$.
A smaller reconstruction loss value intuitively means the reconstructed image is more similar to the original image at a pixel level. In other words, $\tilde{I}_{i}^{m}$ is the synthesized image at the same age as the identity reference image $I_{i}^{m}$. Furthermore, if the reconstruction loss value is zero, $\tilde{I}_{i}^{m}$ is $I_{i}^{m}$ exactly. Since we do not have any identity information about training data, the original image is used as the ground truth for adversarial training. 

Following \cite{zhang2017age}, the identity feature is assumed to follow a uniform distribution, so the adversarial process forces the estimated identity manifold covering the identity distribution as best as possible. Denoting with $p_{data}(I)$ the distribution of the identity reference data $I$ and $p_z$ the prior uniform distribution of identity feature $z_I$, the identity feature is trained to approximate a uniform distribution by:
\begin{align} \label{eq:Dz}
\mathcal{L}_{z_I}=\min_{E_I}\max_{D_I}&~\mathbb{E}_{z_I}{\sim{p_z}}log[D_I(z_I)]+\mathbb{E}_{I\sim{p_{data}(I)}}log[1-D_I(E_I(I))],
\end{align}
where $E_I(\cdot)$ is the identity feature, and $E_{I}$ and $D_{I}$ denote the identity encoder and identity discriminator.

{\bf Identity preservation:}
 To further guarantee that the synthesized images preserve the identity information, we introduce the identity preservation loss into the identity agent:
 \begin{align} \label{eq:ID_loss}
\mathcal{L}_{id}=||E_I(I_{i}^{m})-E_I(\tilde{I}_{i}^{n})||_2.
\end{align}
The identity preservation loss enhances the identity feature learning ability. Synthesized images of the same identity at different ages are given the same identity information by minimizing $\mathcal{L}_{id}$, which disentangles the identity feature from the age feature.

\subsection{Generator and Discriminator}

Following the work in \cite{zhang2017age}, the generator $G$ and the image discriminator $D$ have the same architecture as CAAE, except the input: the input of DRAS consists of two images, one is for identity reference and the other is for age reference, while CAAE requires an image for identity reference and a number for age reference.

To generate a photo-realistic face image, the discriminator tries to discriminate the two reference images as real and the generated image as fake. Thus, the adversarial loss function can be derived as:
\begin{align} \label{eq:Dimg}
\mathcal{L}_{adv}=\min_{G}\max_{D}~&\mathbb{E}_{I_{i}^{m}\sim{p_{data}(I)}}log[D(I_{i}^{m})]+\mathbb{E}_{I_{j}^{n}\sim{p_{data}(I)}}log[D(I_{j}^{n})]+\mathbb{E}_{I_{i}^{m},I_{j}^{n}\sim{p_{data}(I)}}log[1-D(\tilde{I}_{i}^{n})],
\end{align}
where $\tilde{I}_{i}^{n}=G(E_I({I}_{i}^{m}),E_A({I}_{j}^{n}))$. As with the original GANs, the generator and discriminator are alternately optimized via the adversarial loss.

\subsection{Objective Function}

To guarantee the performance of our model, a hybrid loss function is constructed, which consists of the identity feature preservation loss, the age feature preservation loss and two adversarial losses. Equation \eqref{eq:total_loss} shows the overall objective function:
\begin{align}\label{eq:total_loss}
\min_{E_I,E_A,G}  \max_{D_I,D}&~\lambda_{adv}\mathcal{L}_{adv}+\lambda_{id}(\mathcal{L}_{z_{I}}+\mathcal{L}_{rec}+\mathcal{L}_{id})+\lambda_{age}\mathcal{L}_{age},
\end{align}
where $\lambda_{adv}$, $\lambda_{id}$ and $\lambda_{age}$ are weights to control the impact of these loss terms. 

The identity agent, the age agent and the generator are optimized by minimizing Equation~\eqref{eq:total_loss}, and the discriminators are optimized by maximizing Equation~\eqref{eq:total_loss}.

\section{Experiments}

\subsection{Data Description}
\begin{figure*}[pos=htbp] 
\centering 
\begin{minipage}[c]{0.4\textwidth} 
\centering
\centerline{\includegraphics[scale=0.3]{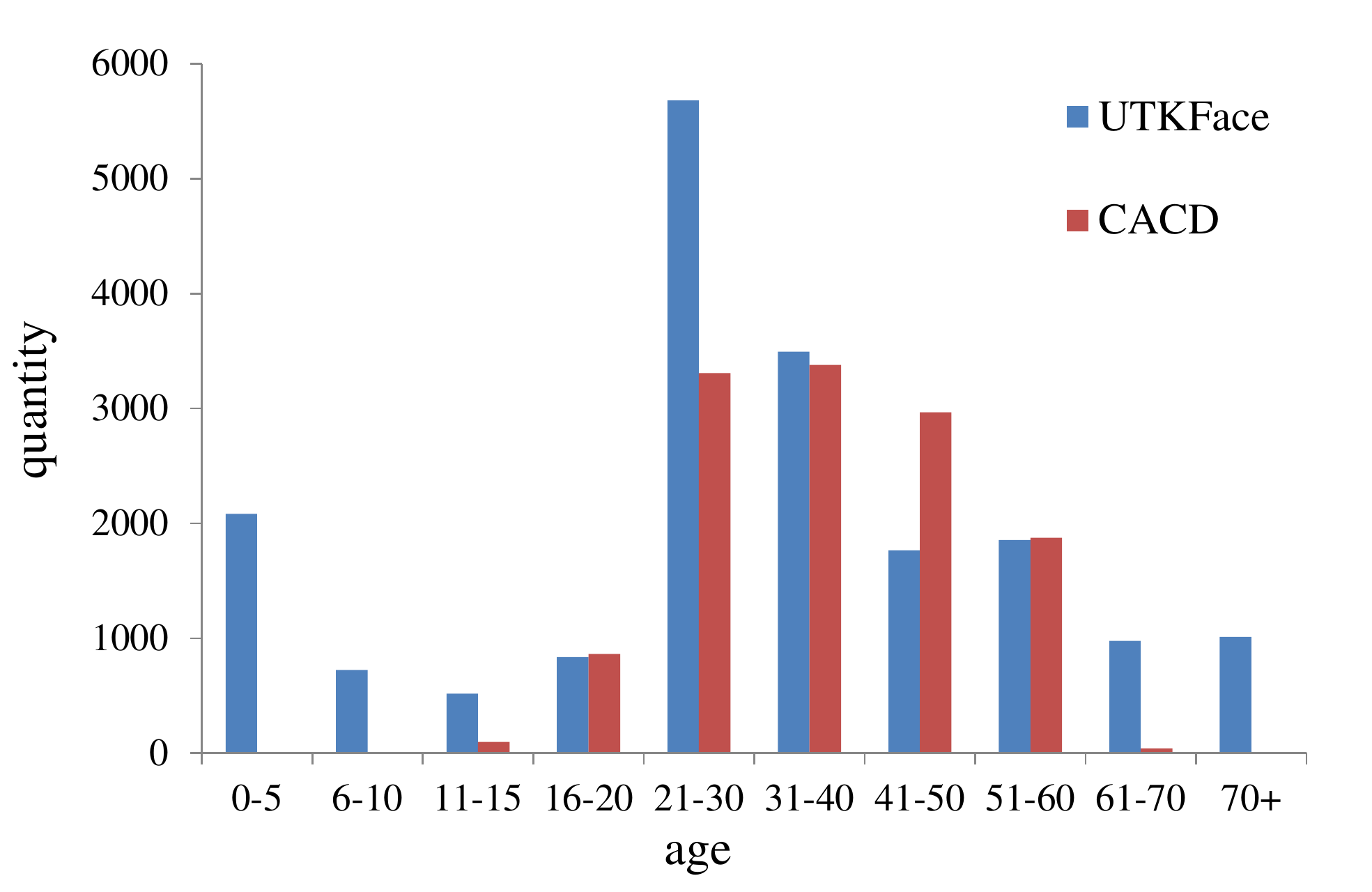}} 
\caption{Original data distributions of CACD \protect\\and UTKFace.} 
\label{fig:agedistribution} 
\end{minipage} 
\hfill
\begin{minipage}[c]{0.5\textwidth} 
\centering 
\centerline{\includegraphics[scale=.25]{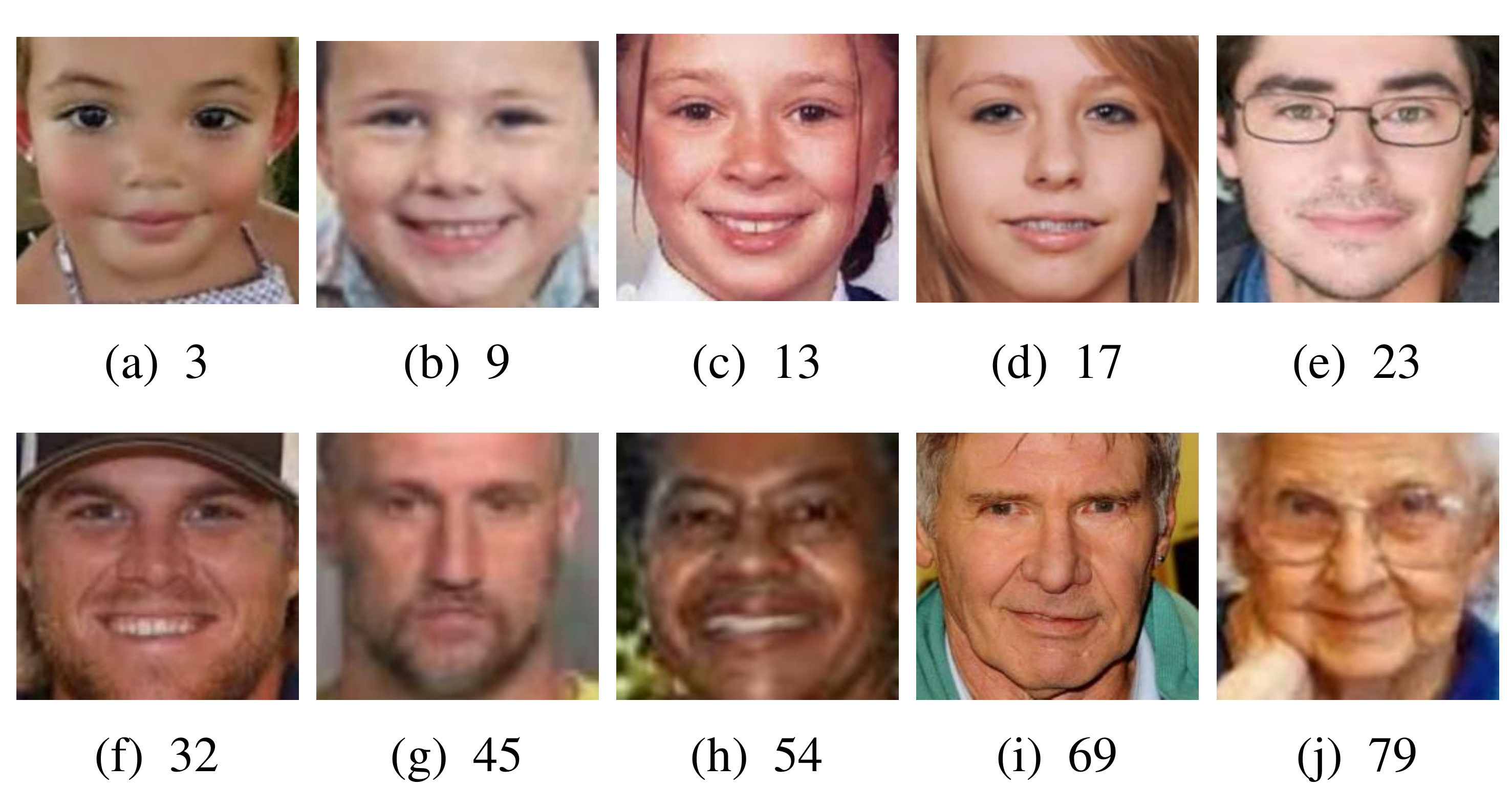}} 
\caption{Ten age reference images. The numbers under \protect\\ each age reference image are the age annotations.}
\label{fig:refimg}
\end{minipage} 
\end{figure*}

We conduct experiments on two widely used benchmark face datasets: UTKFace \cite{zhang2017age}\footnote {https://susanqq.github.io/UTKFace/} and Cross-Age Celebrity Dataset (CACD) \cite{chen2014cross}\footnote {http://bcsiriuschen.github.io/CARC/}. There are over 20,000 facial images without identity annotations in the UTKFace dataset, and 2,000 celebrities in the CACD dataset. Note that though images in UTKFace are in-the-wild, most images are of good quality. However, images in CACD with rank higher than five are ``low quality'', for example, some have wrong identity labels or wrong age labels which can't be used to verify the abilities of identity preservation and age preservation, and some are even not photoes of real person that can't be the reference image of our model. Therefore, we choose those images with rank smaller or equal to five \cite{chen2014cross}. Images are divided into ten age groups according to their age annotations (real ages). Figure \ref{fig:agedistribution} shows the age distributions. Only UTKFace includes babies (zero to five-years-old), children(six to ten-years-old) and senior people (above 70-years-old). The number of people between 20-years-old and 40-years-old is about as twice that of other age groups. In terms of morphology, children (under ten-years-old) have different facial appearances from teenagers and adults, \textit{e.g.} different width between their eyes, face shapes, \textit{etc.} In order to avoid over-fitting or under-fitting, we augment UTKFace and CACD by flipping images of babies, children and seniors.

\subsection{Implementation Details}

80$\%$ of images are used as training data, 10$\%$ as validation data and the remaining $10\%$ as test data. All images are aligned and cropped, and normalized to [-1,1]. And the identity and age features are also normalized to [-1,1] to be unified with the reference images. We train our model on an NVIDIA TITAN X GPU with a decreasing learning rate (the default learning rate is $2e^-3$). We use a mini-batch size as 100, and set $\lambda_{adv}$ as 1, $\lambda_{id}$ as $1e^-3$ and $\lambda_{age}$ as $1e^-2$.

Different from other typical methods, such as \cite{song2018dual}, the DRAS takes two images as inputs and doesn't need any annotations of the identity or the age. Therefore, in the training step, we choose one image as the identity reference image and another as the age reference image randomly. It worth noting that both the two reference images are sampled randomly from the same training dataset, which means that an identity reference image can also be an age reference image and vice versa. Therefore, the reconstruction loss $\mathcal{L}_{rec}$ and the adversarial loss $\mathcal{L}_{Z_I}$ are only related to the identity reference image. Furthermore, our model will still work if the identity reference image is replaced by the age reference image in these two loss functions.    

Empirically, it is difficult to achieve good performance if we train the model with the hybrid loss function directly. Thus, we apply a joint-training strategy. First, in order to learn the age and identity information and ensure that the approximated identity manifold covers the whole feature space, we set $I_{i}^{m}=I_{j}^{n}$ to reconstruct $I_{i}^{m}$. In reconstruction stage, the identity agent is trained with the reconstruction loss $\mathcal{L}_{rec}$ and the adversarial loss $\mathcal{L}_{z_{I}}$, and the age agent is trained with the reconstruction loss $\mathcal{L}_{rec}$ and the age preservation loss $\mathcal{L}_{age}$. Furthermore, to guarantee the generated images be photo-realistic, the discriminator $D$ and the generator $G$ are trained with the other adversarial loss $\mathcal{L}_{adv}$ alternatively. Subsequently, after the losses of the identity agent and the age agent  converge, we fix $E_I$, $E_A$ and $D_I$, set $I_{i}^{m}\neq{I_{j}^{n}}$, and use the two preservation functions $\mathcal{L}_{id}$ and $\mathcal{L}_{age}$ to optimize the generator and discriminator.

\subsection{Experimental Performance and Analysis}

In this section, we first investigate the performance of our model, then select two baselines, CAAE \cite{zhang2017age} and IPCGAN \cite{wang2018face}, for comparison. Since conventional GAN-based methods use ten age groups to investigate different age effects, for fair comparison, we randomly choose one image from each group as the age reference images. Note that these ten age reference images are distinct from the training images to avoid over-fitting, as shown in Figure \ref{fig:refimg}. As can be seen, the men in Figure \ref{fig:refimg}(f), (g) and (h) are from different age groups according to their age annotations, but they look as old as each other.

\subsubsection{Performance Evaluation of Disentangled Identity Feature Learning}
Disentangled identity feature here means the identity features of different people should be isolated from each other, regardless of whether or not they are in the same age group. T-Distributed Stochastic Neighbor Embedding(t-SNE) \cite{maaten2008visualizing,van2012visualizing} depicts the similarities of identity features, which can be used to visualize the disentangled their disentangled representations. The t-SNE model outputs similar identity features for nearby points and dissimilar ones for distant points. Since the data in CACD have identity annotations, we evaluate the disentangle feature learning performance on this dataset. Images of 11 celebrities with different ages are collected from CACD, as described in Table \ref{tab:id_fea}. 

\begin{table}
\parbox{.45\linewidth}{
\centering
\caption{Description of 11 subsets for identity feature \protect\\ learning performance analysis.}
\label{tab:id_fea}
\begin{tabular}{cc}
\hline
Identity & Age Range \\
\hline
id0 & 14-23 \\
id1 & 15-23 \\
id2 & 15-24\\
id3 & 17-26\\
id4 & 20-29\\
id5 & 21-30\\
id6 & 26-33\\
id7 & 32-41\\
id8 & 43-52\\
id9 & 46-55\\
id10 & 49-58\\
\hline
\end{tabular}
}
\hfill
\parbox{.45\linewidth}{
\centering
\caption{Description of four ablation models.}
\label{tab:test scenarios}
\begin{tabular}{cc}
\hline
Model & Description \\
\hline
M1 & with the two preservation functions \\
M2 & without the two preservation functions \\
M3 & with only the identity preservation function\\
M4 & with only the age preservation function\\
\hline
\end{tabular}
}
\end{table}

\begin{figure}[pos=h]
    \centering
    \includegraphics[scale=0.35]{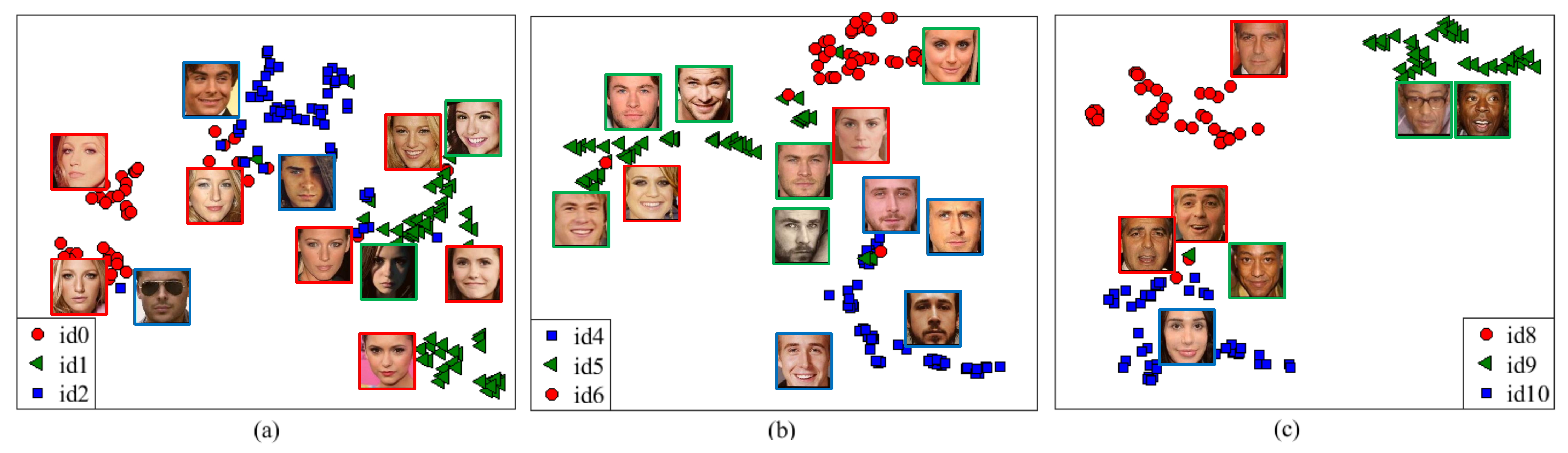}
    \caption{Visualization of identity features. (a) Age ranges from 14 to 26 years old; (b)Age ranges from 20 to 33 years old; (c) Age ranges from 43 to 58 years old.}
    \label{fig:tsne-DRAS-1new}
\end{figure}

Nine subsets from Table \ref{tab:id_fea} are chosen and divided into three groups \{[$id0$, $id1$, $id2$], [$id4$, $id5$, $id6$], [$id8$, $id9$, $id10$]\} at the same age respectively. To examine the performance of disentangled identity feature learning on the three groups, identity features of the nine people are retrieved using the identity agent for visualization, shown in Figure \ref{fig:tsne-DRAS-1new}. The nine people are almost entirely isolated from each other. However, since pose, expression, face shape, \textit{etc.} represent identity feature, some people with simialr poses, expressions or face shape overlap. For example, $id0$ and $id1$ have profiles in Figure \ref{fig:tsne-DRAS-1new}(a), $id5$ and $id6$ have similar smiling and serious expressions in Figure \ref{fig:tsne-DRAS-1new}(b)), and $id8$ and $id10$ both have long face shapes in Figure \ref{fig:tsne-DRAS-1new}(c)).  

\begin{figure}[pos=h]
    \centering
    \includegraphics[scale=0.35]{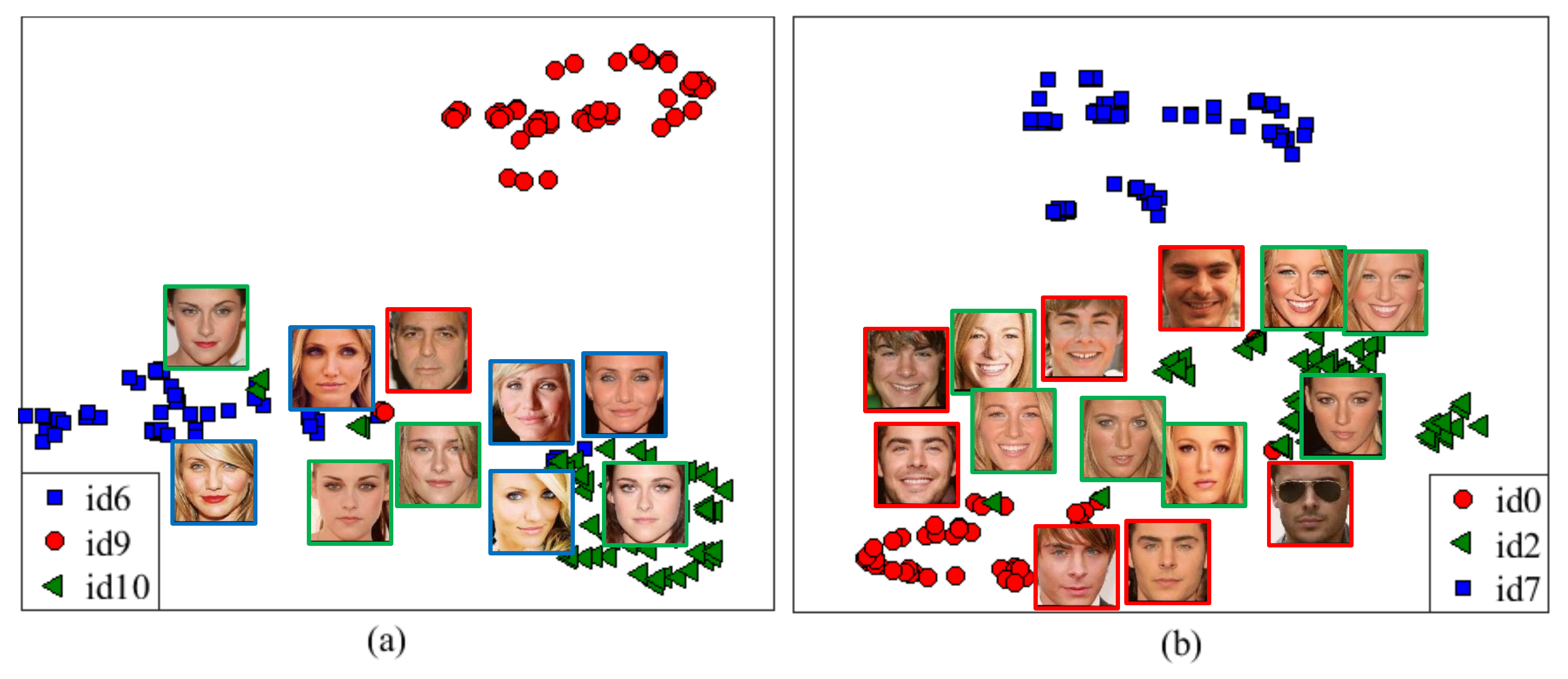}
    \caption{Visualization of identity features in different age ranges. (a) $id0$, $id9$ and $id10$ are from three different age ranges; (b) $id0$ and $id2$ are from the same age range, and $id7$ is from another age range.}
    \label{fig:tsne-DRAS-2new}
\end{figure}

Moreover, we randomly select six individuals from Table \ref{tab:id_fea} to study the identity features of different people at various ages. In Figure \ref{fig:tsne-DRAS-2new}, we can find that most samples are correctly isolated in the identity feature space. However, these are still some overlapping points across different identities. To further explore this, we plot out the corresponding images of those overlapping points. It can be seen that some of overlapping are caused by similar makeups, \textit{e.g.} $id6$ and $id10$, and some are because of their sharing similar expressions, \textit{e.g.} $id9$ and $id10$. It is also interesting to note that, these overlapping points prove the manifold assumption from the side.

In these two experiments, most of the identity features of different people fall in different clusters, regardless whether or not they are the same age, validating the disentangled feature learning ability of the identity agent.

\subsubsection{Ablation Study}\label{sec:ablation}

\begin{figure}[pos=h]
    \centering
    \includegraphics[scale=0.3]{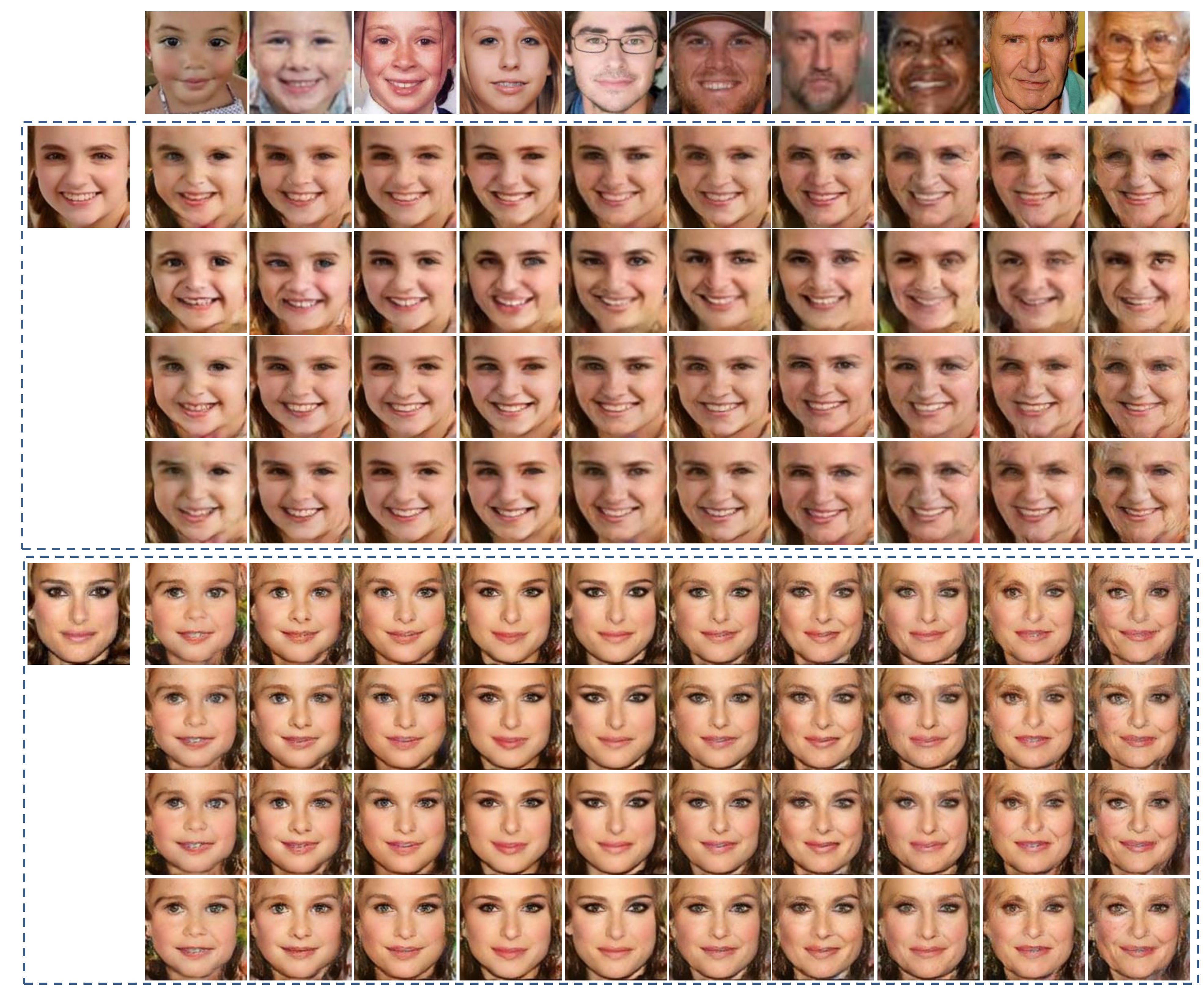}
    \caption{Effects of the identity and age preservation functions. From top to bottom, facial images are generated by $M1$, $M2$, $M3$ and $M4$, respectively.}
    \label{fig:Eff_IDAGE}
\end{figure}

To validate the effects of the identity and age preservation losses, we design four ablation models, abbreviated $M1$, $M2$, $M3$ and $M4$, as shown in Table \ref{tab:test scenarios}. The images generated by the four models are shown in Figure \ref{fig:Eff_IDAGE}. 

\begin{figure}[pos=h]
    \centering
    \includegraphics[scale=0.3]{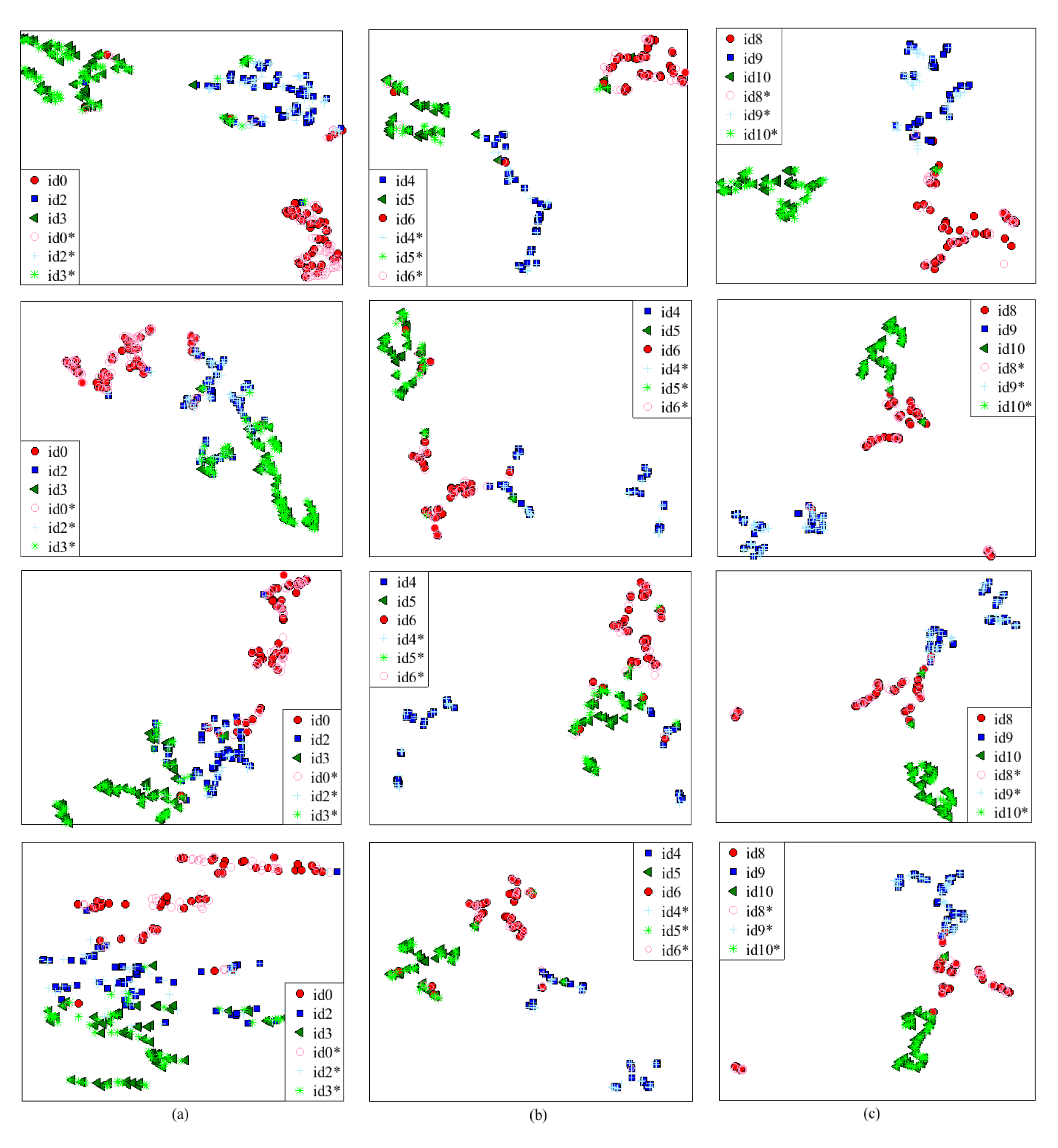}
    \caption{Effects of identity preservation function. From top to bottom: identity feature visualizations obtained under $M1$, $M2$, $M3$ and $M4$, respectively. $idn*$ is the synthesized image for identity reference image $idn$. (a) 14-26 years old; (b) 20-33 years old; (c) 43-58 years old.}
    \label{fig:tsne-ablation-1}
\end{figure}

\begin{figure}[pos=h]
    \centering
    \includegraphics[scale=0.3]{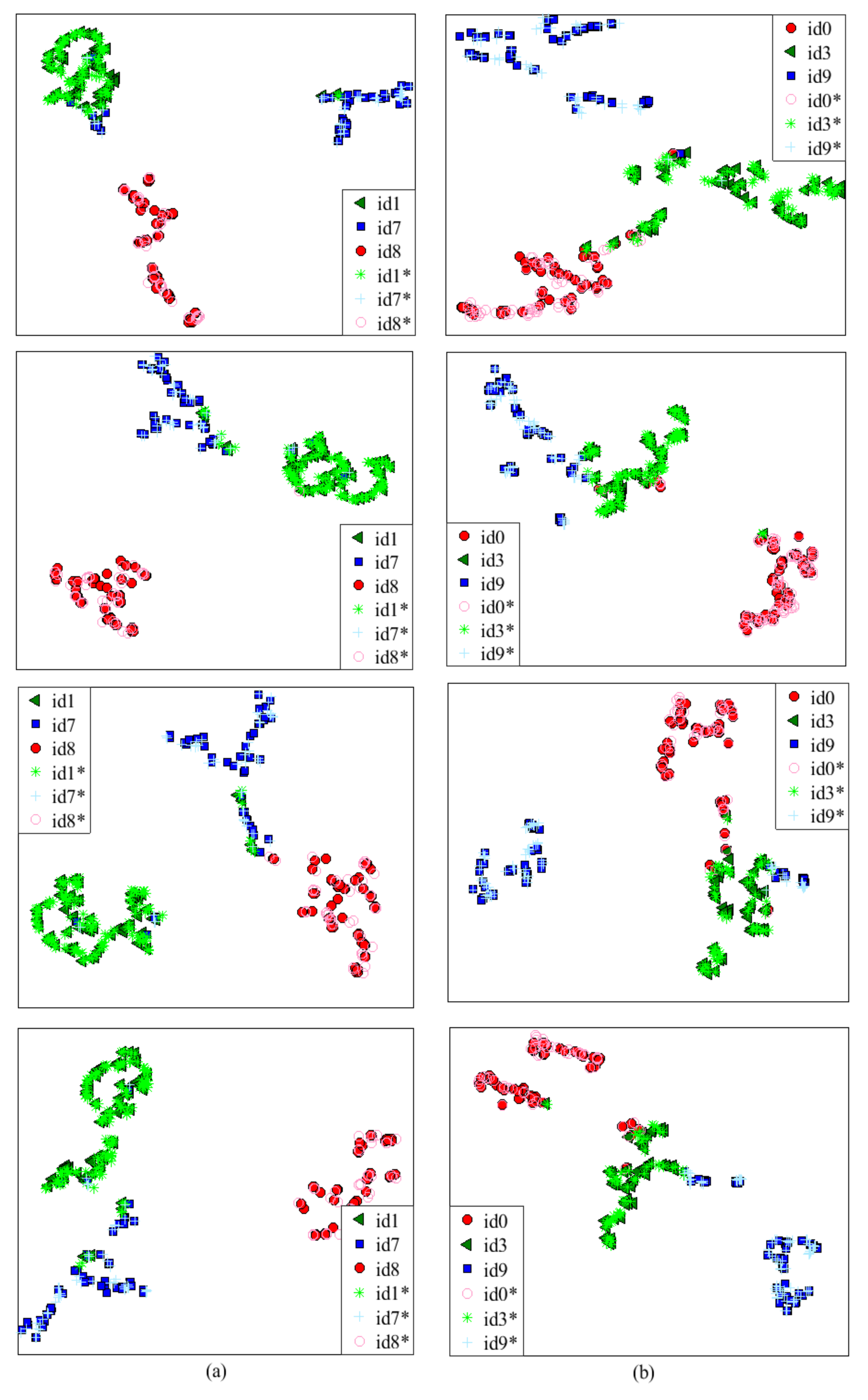}
    \caption{Effects of identity preservation loss. From top to bottom: visualizations of identity features obtained under $M1$, $M2$, $M3$ and $M4$, respectively. $idn*$ is the synthesized image for identity reference image $idn$. (a) $id1$, $id7$ and $id8$ are from three different age ranges; (b) $id0$ and $id3$ are from the same age range, and $id9$ is from another age range.}
    \label{fig:tsne-ablation-2}
\end{figure}

In Figure \ref{fig:Eff_IDAGE}, some images generated by $M2$ and $M3$ look younger than their age reference images, some images of $M2$ look male, but are actually female, and some images of $M4$ have artifacts on the local facial parts. For the younger appearances of $M2$ and $M3$, it is mainly caused by the lack of the age preservation loss. For the incorrect male appearance of $M2$, it is mainly caused by the lack of the identity preservation loss. For the undesired artifacts of $M4$, it is due to only considering the age preservation loss, which promotes to generate images with more age information related features, such as wrinkles in the eye corner or wide eyes in children \textit{etc.}, which can be seen as artifacts by human eyes.

{\bf Effect of Identity Preservation}
To intuitively evaluate the effect of the identity preservation loss in the feature space, we also use t-SNE to visualize the identity features of the identity reference images and the generated images. Figure \ref{fig:tsne-ablation-1} shows the identity features of different people of the same age and Figure \ref{fig:tsne-ablation-2} shows the identity features of different people at different ages. Images in the first row ($M1$) have the best performance in terms of in-cluster compactness and between-cluster separation, which demonstrates that the two preservation functions in DRAS can effectively conserve identity features and perform disentangling. Note that, the generated images have slight feature shifts from their identity reference images, which is caused by the compromise between the identity and age preservation losses. There are more overlapping points and more feature shifting in the second and the fourth rows ($M2$ and $M4$), caused by the lack of identity preservation loss. In the third row ($M3$), the identity features of the generated images tend to overlap with their identity reference images, which validates the fact that the identity preservation loss works in conjunction with the identity feature space.

\begin{figure}[pos=h]
    \centering
    \includegraphics[scale=0.2]{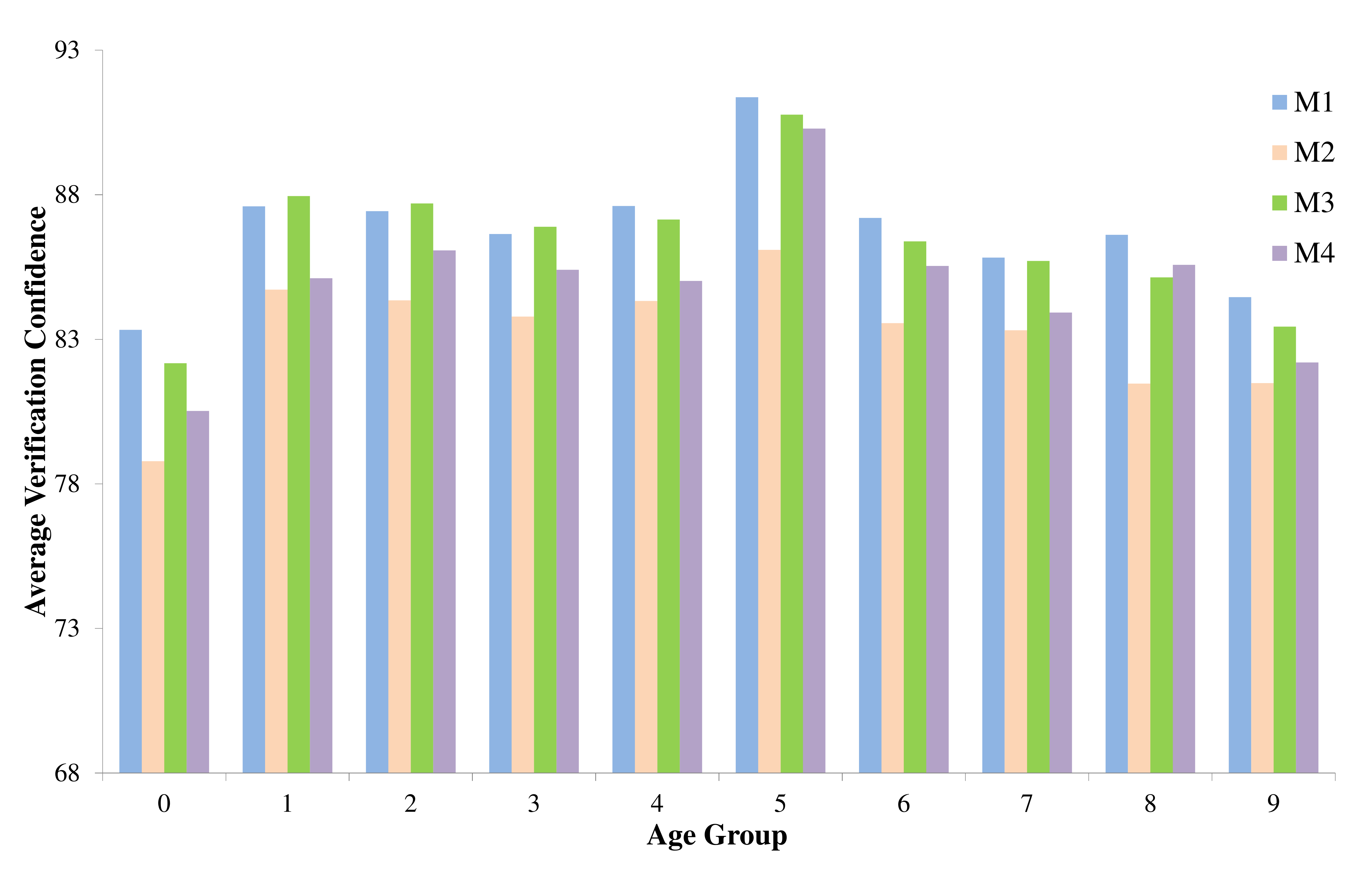}
    \caption{Quantified identity consistency of different age groups with 4 models. Y-axis shows the average verification confidence between age group $i$ with age groups $j$. }
    \label{fig:ID_ABLATION}
\end{figure}

We also use the online face comparator provided by Face++ to quantify the effect of identity preservation loss \cite{face_comp}. The confidence threshold is set as 73.975, and higher confidence  means large likelihood between two face images. First, we compare each test image with it's corresponding synthesized images.  The comparative results in Table \ref{tab:ablation_id} indicate those models with the identity preservation function, \textit{i.e.} model $M1$ and $M3$, generally obtain higher verification confidence. And the quantitative result of $M1$ is lower than that of $M3$ which also demonstrates the compromise between the identity and age preservation losses, for example the wrinkle on face is not clear and makes face looked dirty. Note, the verification confidence gaps among the 4 models are small which explain the reconstruction loss plays an important role in identity perservation. Furthermore, to investigate the consistency of identity, we divide the synthesized images into 10 age groups and conduct the comparision among these age groups, \textit{i.e.} [age group $i$, age group $j$] ($i \neq j$). The average confidences in Figure \ref{fig:ID_ABLATION} show model $M1$ retain more identity consistency along with face aging.     

\begin{table}
\centering
\caption{Comparative results of identity preservation with four models. Model $M1$ and $M3$ generally obtain higher verification confidence.}
\label{tab:ablation_id}
\begin{tabular}{cc}
\hline
Model & Average Verification Confidence \\
\hline
M1 & 80.37$\pm$3.13 \\
M2 & 78.98$\pm$0.63 \\
M3 & 82.10$\pm$2.83 \\
M4 & 79.19$\pm$1.71\\
\hline
\end{tabular}
\end{table}

By visual and quantitative evaluation, we conclude that the identity preservation function does enhance the model ability of identity preserving and consistency.

{\bf Effect of Age Preservation}
To discern whether a synthesized image has the same age feature as the age reference image, we use a pre-trained AlexNet model fine-tuned on UTKFace and CACD. Since each test image has ten synthesized images, the data is balanced. Additionally, we can use accuracy to describe age preservation performance: a higher accuracy suggests that more generated images have the same age features as their age reference images. 

Age similarities in the $M1$, $M2$, $M3$ and $M4$ models are measured and the results are shown in Table \ref{tab:eff_age}. DRAS with age preservation function ($M1$ and $M4$) obtains higher accuracy than that without it. Images are synthesized by model $M1$ using both the age preservation loss and the identity preservation loss, which leads to a compromise between accuracy in identity and age. Therefore, the average performance of DRAS using only the age preservation function is the highest. 

\begin{table}
\centering
\caption{Effect of age preservation function. Bold numbers are the maximums.}
\label{tab:eff_age}
\begin{tabular}{m{2.4cm}<{\centering}m{2.6cm}<{\centering}m{2.6cm}<{\centering}m{2.6cm}<{\centering}m{2.6cm}<{\centering}}
\hline
Age Groups & Accuracy of $M1$($\%)$ & Accuracy of $M2$($\%)$ & Accuracy of $M3$($\%)$ & Accuracy of $M4$($\%)$\\
\hline
0$\#$ (0-5) &\textbf{99.93} & 99.8 & 99.86 & \textbf{99.93}\\
1$\#$ (6-10) & \textbf{99.63} & 92.85 & 93.32 & 99.49 \\
2$\#$ (11-15)  & 92.47 & 74.44 & 74.88 & \textbf{93.86}\\
3$\#$ (16-20)  & \textbf{93.12} & 78.98 & 80.41 & 92.17\\
4$\#$ (21-30)  & \textbf{100.00} & 99.97 & 99.9 & \textbf{100.00}\\
5$\#$ (31-40)  & \textbf{93.90} & 89.97 & 90.54 & 93.83\\
6$\#$ (41-50)  & 84.71 & 71.59 & 74.1 & \textbf{85.12}\\
7$\#$ (51-60) & 90.10 & 92.51 & \textbf{97.63} & 97.02\\
8$\#$ (61-70)  & 97.29 & 98.20 & 99.53 & \textbf{99.66}\\
9$\#$ (70+)  & \textbf{99.97} & 98.58 & 99.56 & 99.90\\
\hline
Average Acc.($\%$)  & 96.04 & 89.36 & 90.33 & \textbf{96.15}\\
\hline
\end{tabular}
\end{table} 

\subsubsection{Generative Performance Comparison}\label{sec:G-COM}
In this experiment, the performances of DRAS, CAAE and IPCGAN are compared in terms of their generated images. We use the codes published online by the authors\footnote{https://github.com/ZZUTK/Face-Aging-CAAE}\footnote{https://github.com/dawei6875797/Face-Aging-with-Identity-Preserved-Conditional-Generative-Adversarial-Networks}, with the same configurations as the original papers. For fair comparisons with the two baselines, we take the following experimental protocol: for the baseline CAAE, since its released model is trained on UTKFace, we replicate the experimental results of UTKFace and fine-tune the model on CACD; for the other baseline, IPCGAN, since its released model is trained on CACD, we replicate the experimental results of CACD and fine-tune the model on UTKFace. In this paper, as the released codes of IPCGAN were trained to synthesize images in five age groups (11-20, 21-30, 31-40, 41-50 and 50+), we use the same categories to generate images.

\begin{figure}[pos=h]
    \centering
    \includegraphics[scale=0.3]{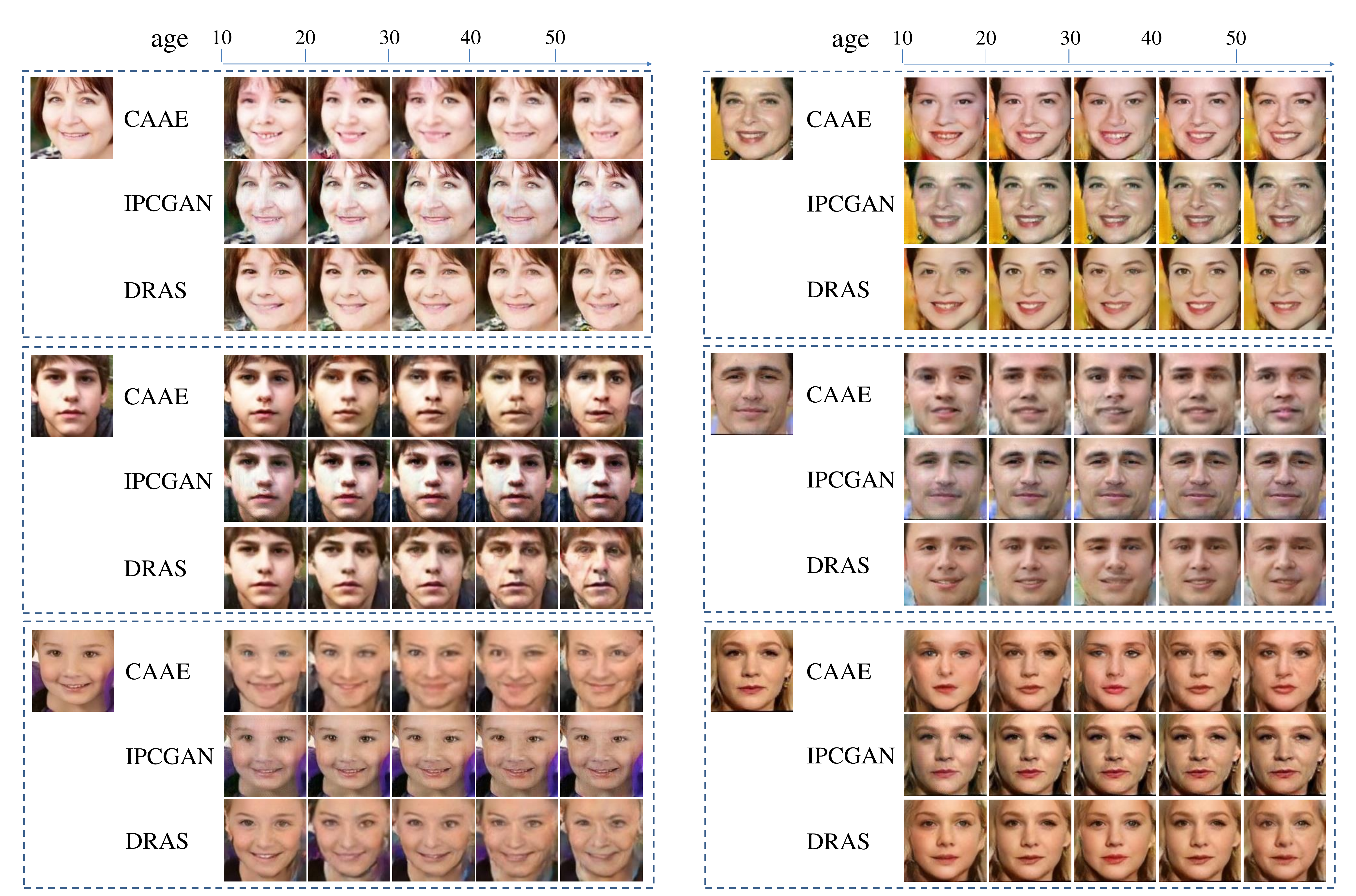}
    \caption{Some synthesized faces for UTKFace and CACD. Each dotted box denotes images of the same person. The first column on the left in each dotted box is the identity reference image for DRAS (the input images for CAAE and IPCGAN).}
    \label{fig:COMP43}
\end{figure}

First, images are generated by the three respective methods, taking Figure \ref{fig:refimg}(d)-(h) as their age reference images. In Figure \ref{fig:COMP43}, from left to right, the facial images generated by DRAS get older as their age reference images do. The generated images retain the age features of the reference images, \textit{e.g.} round cheeks and bigger eyes for younger images. Regarding pose and expression, the generated images have the same identity features as their identity reference images. From Figure \ref{fig:COMP43}, we can also see that the age effects of the images generated by IPCGAN change slightly, while the synthesized images of CAAE look blurry and have artifacts. For IPCGAN, the slight aging effect is caused by the fact that it cannot effectively isolate the identity and age features from each other, since the identity features share the same part of convolutional network as the age classifier. For CAAE, the undesirable artifacts are inevitable because it only uses the reconstruction loss to preserve identity features without constraining the age preservation. Thus, compared with CAAE and IPCGAN, our model can generate higher quality facial images whose identities and ages are consistent with their reference images.

\begin{figure}[pos=h]
    \centering
    \includegraphics[scale=0.3]{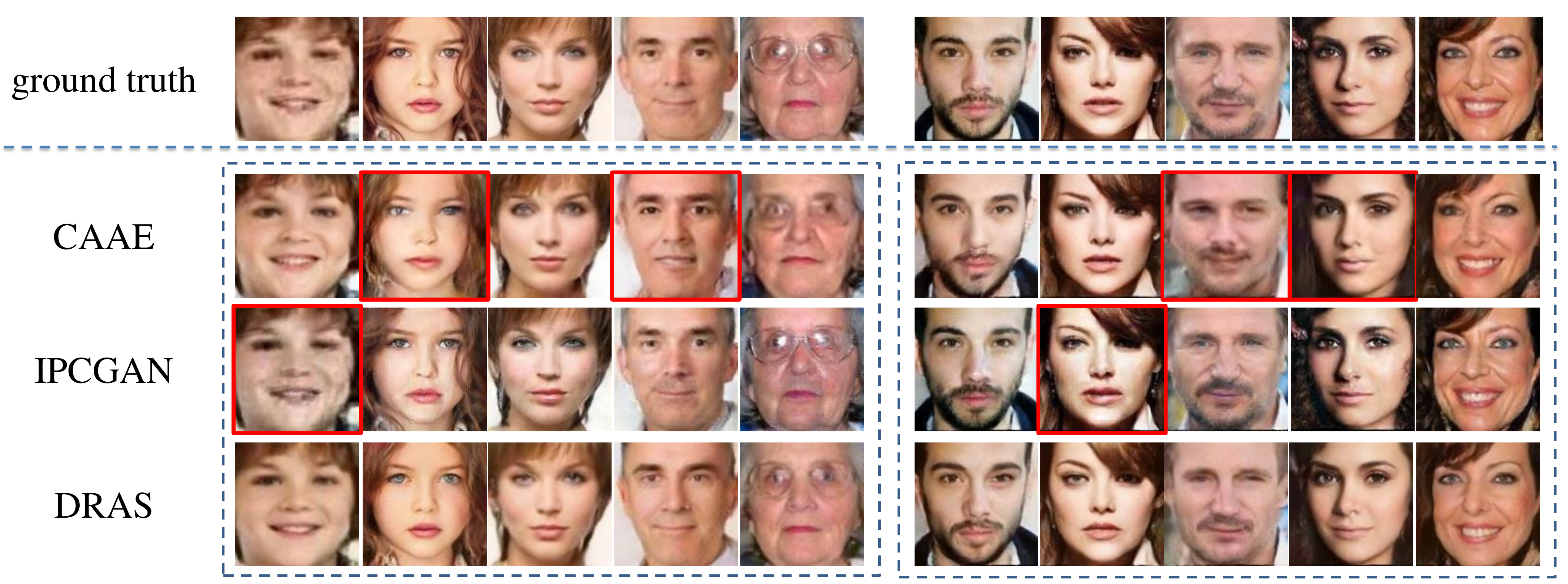}\caption{Reconstruction results of the CAAE, IPCGAN and DRAS. Images in the left dotted box are from UTKFace and images in the right dotted box are from CACD. Images with red boxes are different from their ground truth or blurry in local facial regions.}\label{fig:REC}
\end{figure}

 The images generated for the five age groups do not have corresponding ground truths. However, if both the identity and age information come from the identity reference image, then the image is equivalent to the ground truth. Therefore, for the dual reference images, we use one image as both the identity and age reference. The generative performance can also be evaluated by comparing the generated images with their ground truth, which ideally should be the same. As can be seen in Figure \ref{fig:REC}, the images generated by DRAS look similar to or even clearer than their ground truth. For CAAE and IPCGAN, images with red boxes are those that look different from their ground truths or are blurry in local facial parts. 

\begin{figure}[pos=!ht]
    \centering
    \includegraphics[scale=0.3]{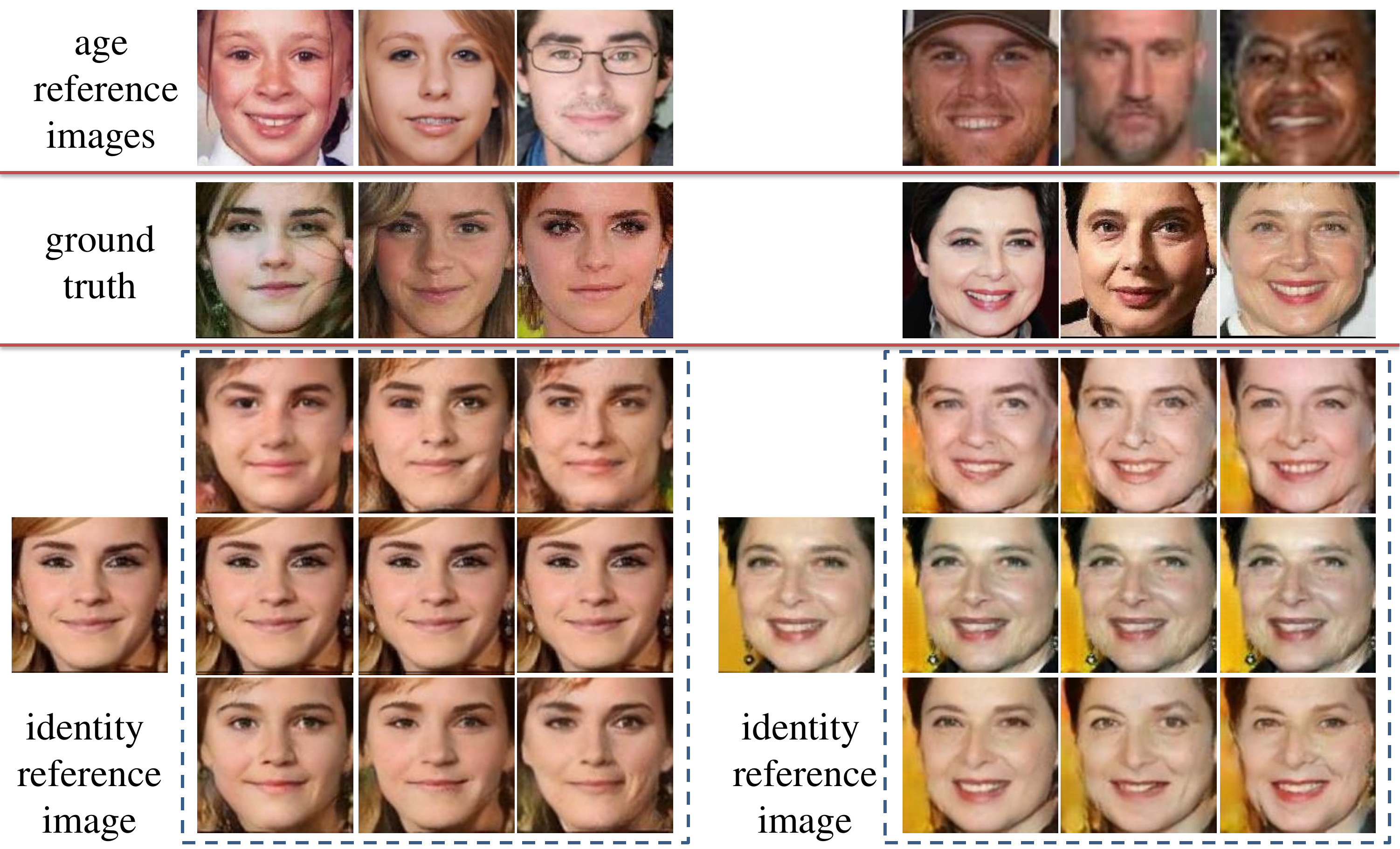}
    \caption{Synthesized faces with different identity reference images and age reference images. The first row shows the age reference images, while the first column on the left shows the identity reference images. The first, second and third row in each dotted box show the results for CAAE, IPCGAN and DRAS, respectively.}
    \label{fig:VSGT}
\end{figure}

Furthermore, facial images for Emma Waston and Isabella Rossellini were generated using different age reference images. Real images the same age as the reference images were taken as the ground truth. As shown in Figure \ref{fig:VSGT}, the results generated by DRAS are most photo-realistic and reasonable. The cheeks in the first synthesized image for Emma Watson just like those of the age reference image. However, the first two synthesized images of Isabella Rossellini look the same age, just as their age reference images. In contrast, images generated by IPCGAN all looked the same as their identity reference images, with no aging effect visible from their faces. For example, from left to right, the first age reference image looks much younger than the third one, yet the third synthesized image of Emma Waston looks just as young as the first. For CAAE, the third synthesized image looks male, which is not acceptable.

Overall, in these three experiments, the generative performance of DRAS is much higher than the other two methods.

\begin{figure}
    \centering
    \includegraphics[scale=0.3]{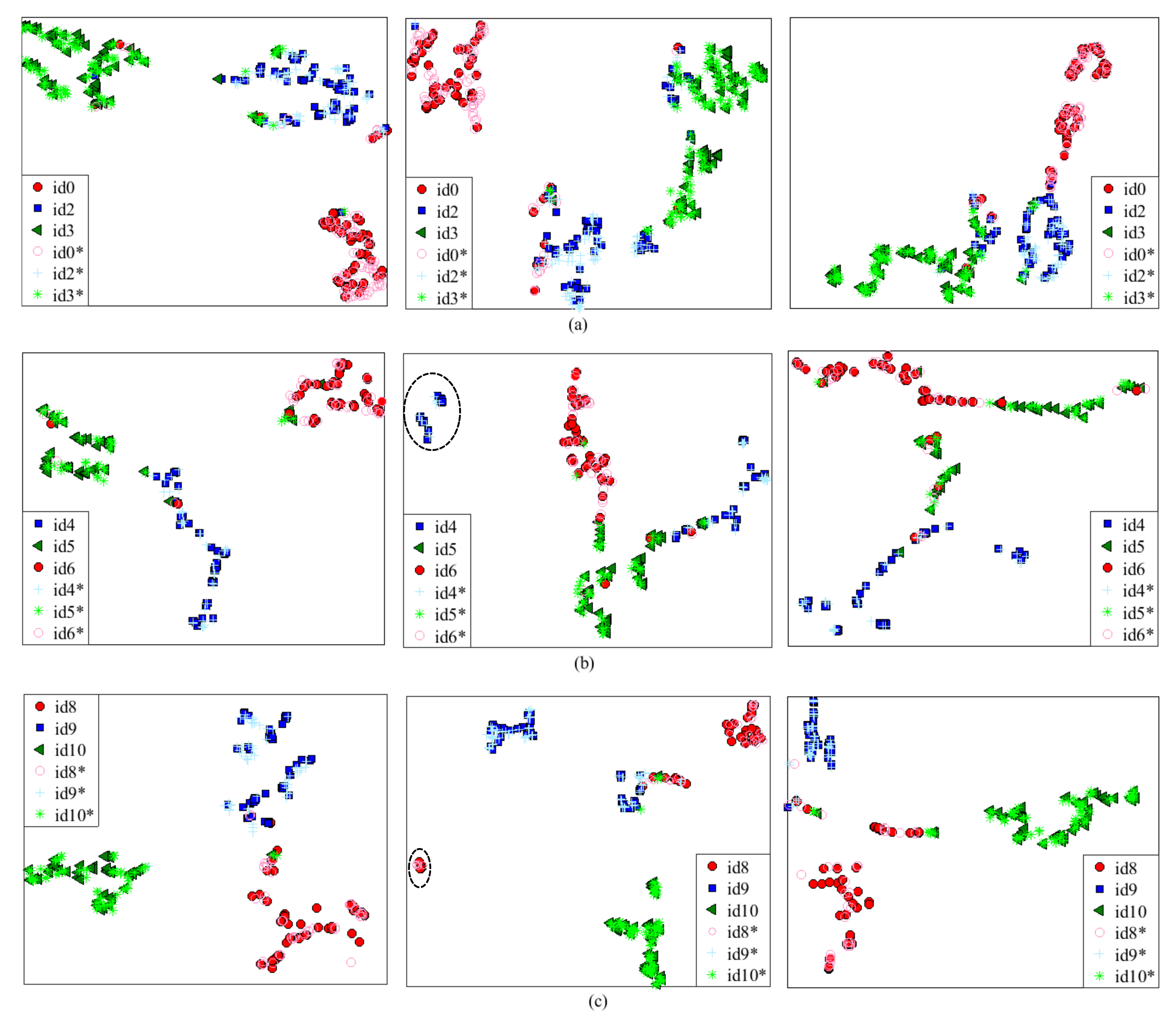}
    \caption{Identity preservation comparison of DRAS, IPCGAN and CAAE. From left to right: they are identity feature visualizations of DRAS, IPCGAN and CAAE, respectively. $idn*$ is the synthesized image of identity reference image $idn$. Points with dotted circles are far away from their feature centers. (a) 14-26 years old; (b) 20-33 years old; (c)43-58 years old.}
    \label{fig:tsne-comp-1}
\end{figure}

\subsubsection{Identity Preservation Comparison}

\begin{figure}[pos=!htb]
    \centering
    \includegraphics[scale=0.3]{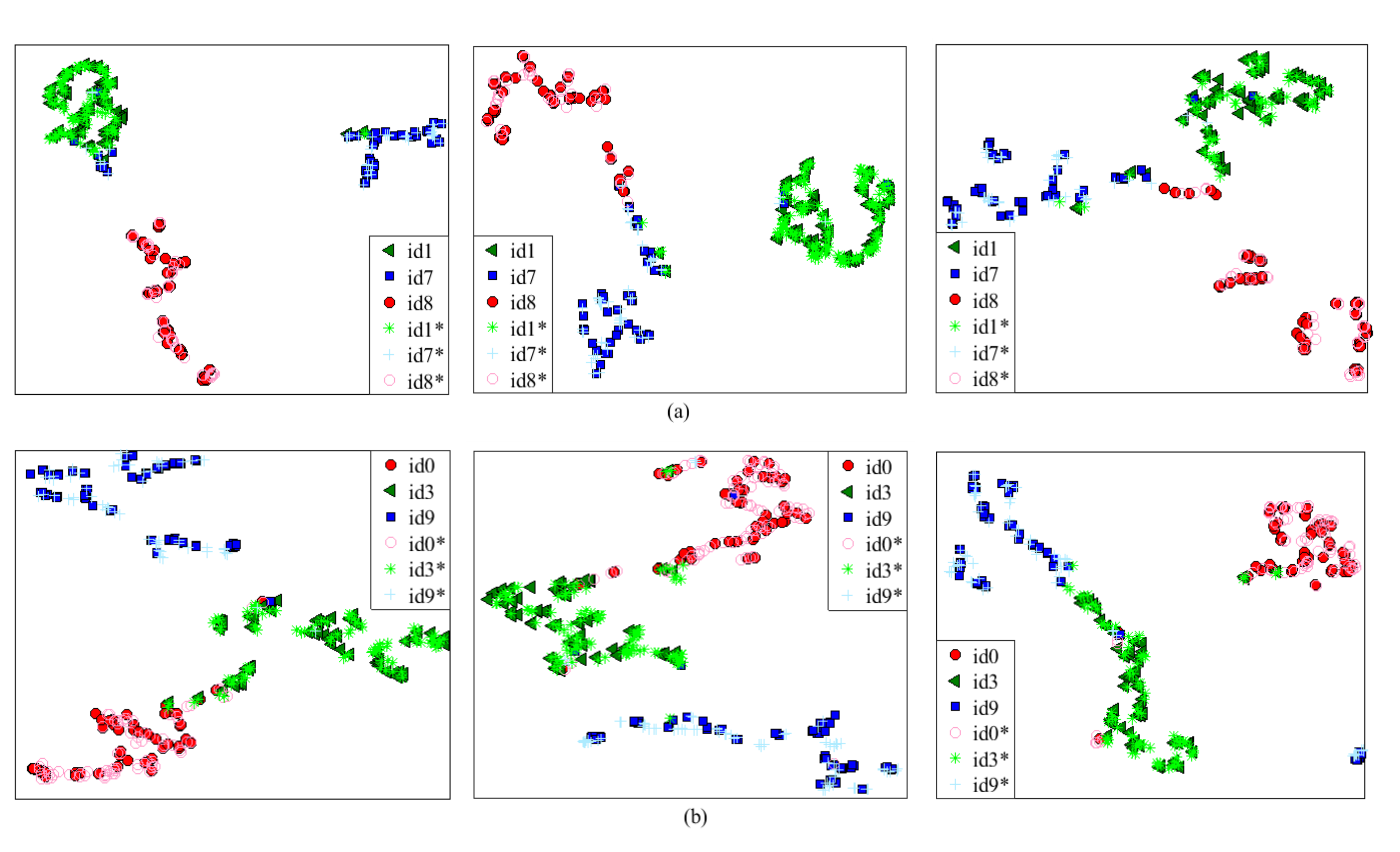}
    \caption{Identity preservation performance of DRAS, IPCGAN and CAAE for different age ranges. From left to right: they are identity feature visualizations of DRAS, IPCGAN and CAAE, respectively. $idn*$ is the synthesized image of identity reference image $idn$. (a) $id1$, $id7$ and $id8$ are from three different age ranges; (b) $id0$ and $id3$ are from the same age range, and $id9$ is from another age range.}
    \label{fig:tsne-comp-2}
\end{figure}

In this experiment, in order to compare the identity preservation capability between our model and the other two methods, t-SNE is again used to visualize the synthesized images in the feature space. Figure \ref{fig:tsne-comp-1} and Figure \ref{fig:tsne-comp-2} show that most synthesized images are close to or even overlapping with their identity reference images in the identity feature space. This is because the three methods all have identity preservation strategy: the identity preservation and reconstruction losses in DRAS, the conceptual and reconstruction losses in IPCGAN, and the reconstruction loss in CAAE. 

The identity features of DRAS and IPCGAN have more intra-cluster compactness, which suggests that only using the reconstruction loss dose not guarantee identity consistency. Moreover, the identity features of DRAS have more inter-cluster separation than the other two methods, demonstrating the disentangled feature learning ability of the identity agent. The t-SNE visualization result of IPCGAN, shown in Figure \ref{fig:tsne-comp-1}, presents several outliers, even for the same person (marked by dotted circles). This is because IPCGAN shares identity feature layers with the age classifier, making it difficult to disentangle the identity and age features. For the visualization result of CAAE, it is clearly undesirable to keep the identity features of different people always joint. CAAE lacks a disentangling ability mainly because it only considers identity preservation on a pixel level rather than in the feature space.

Quantitative face verification between each test image and its synthesized images is carried out to check the identity preservation performances of the 3 methods. The quantitative measurements between the test identity reference image and its synthesized images are shown in Table \ref{tab:comp_id_3methods}. We also perform face verification among synthesized images of the 3 methods and the comparative results of 5 age groups (grp.$n$ for the abbreviation of age group $n$) are shown in Tables \ref{tab:CAAE_id}, \ref{tab:IPCGAN_id} and \ref{tab:DRAS_id}. From Figure \ref{fig:COMP43}, we find that synthesized images of IPCGAN look almost the same as their identity reference images, which is the reason for the highest average verification confidence of IPCGAN. However, it is not an ideal method for lacking of aging effect. For DRAS, the verification confidences in Table \ref{tab:comp_id_3methods} and Table \ref{tab:DRAS_id} are all surpass the threshold and outperform CAAE. As section \ref{sec:ablation} discussed, our model improves the identity preservation ability and renders age effect as well.

\begin{table}
\centering
\caption{Comparative results of identity preservation of the 3 methods. Synthesized images of IPCGAN look almost the same as their identity reference images, which is the reason for the highest average verification confidence of IPCGAN. However, it is not an ideal method for lacking of aging effect.}
\label{tab:comp_id_3methods}
\begin{tabular}{cc}
\hline
Method & Average Verification Confidence \\
\hline
CAAE & 71.43$\pm$2.00 \\
IPCGAN & 94.13$\pm$0.64 \\
DRAS & 81.52$\pm$0.83 \\
\hline
\end{tabular}
\end{table}

\begin{table}
\centering
\caption{Identity consistency of CAAE.}
\label{tab:CAAE_id}
\begin{tabular}{cccccc}
\hline
Age Groups & grp.1 & grp.2 & grp.3 & grp.4 & grp.5\\
\hline
grp.1 & - & 73.53$\pm$7.71 & 71.10$\pm$10.08 & 66.75$\pm$10.40 & 68.696$\pm$6.36\\
grp.2 & - & - & 85.30$\pm$2.51 & 76.65$\pm$4.19 & 72.47$\pm$4.20\\
grp.3 & - & - & - & 79.08$\pm$8.24 & 78.69$\pm$1.80\\
grp.4 & - & - & - & - & 77.036$\pm$2.27\\
\hline
Average Confidence& 75.49$\pm$6.91 & 81.07$\pm$3.72 & 82.31$\pm$4.53 & 79.38$\pm$5.02 & 78.86$\pm$2.93\\
\hline
\end{tabular}
\end{table}

\begin{table}
\centering
\caption{Identity consistency of IPCGAN.}
\label{tab:IPCGAN_id}
\begin{tabular}{cccccc}
\hline
Age Groups & grp.1 & grp.2 & grp.3 & grp.4 & grp.5\\
\hline
grp.1 & - & 95.86$\pm$0.27 & 94.91$\pm$0.16 & 94.65$\pm$0.30 & 94.72$\pm$0.43\\
grp.2 & - & - & 95.78$\pm$0.29 & 95.00$\pm$0.16 & 94.90$\pm$0.50\\
grp.3 & - & - & - & 94.42$\pm$2.17 & 94.25$\pm$1.56\\
grp.4 & - & - & - & - & 96.50$\pm$0.24\\
\hline
Average Confidence& 95.50$\pm$0.22 & 95.78$\pm$0.24 & 95.35$\pm$0.83 & 95.88$\pm$0.58 & 95.55$\pm$0.54\\
\hline
\end{tabular}
\end{table}

\begin{table}
\centering
\caption{Identity consistency of DRAS.}
\label{tab:DRAS_id}
\begin{tabular}{cccccc}
\hline
Age Groups & grp.1 & grp.2 & grp.3 & grp.4 & grp.5\\
\hline
grp.1 & - & 87.09$\pm$4.66 & 89.73$\pm$3.33 & 83.04$\pm$3.39 & 77.27$\pm$8.59\\
grp.2 & - & - & 90.60$\pm$3.38 & 86.90$\pm$2.24 & 82.15$\pm$5.25\\
grp.3 & - & - & - & 89.46$\pm$2.68 & 87.83$\pm$2.85\\
grp.4 & - & - & - & - & 88.21$\pm$3.16\\
\hline
Average Confidence& 86.91$\pm$3.99 & 88.83$\pm$3.11 & 91.00$\pm$2.25 & 89.00$\pm$2.30 & 86.58$\pm$4.00\\
\hline
\end{tabular}
\end{table}

\subsubsection{Age Preservation Comparison}

In the last experiment, we compared the age preserving performance of the three methods. The quantitative comparison results are shown in Table \ref{tab:age_cls} and Table \ref{tab:age_cls_ipcgan}. Our model clearly obtains the highest accuracy for seven age groups, with groups $0\#$, $5\#$ and $7\#$ being the only exceptions, where the performance is slightly lower. When observing the original data distribution in Figure \ref{fig:agedistribution}, the amount of CACD data in group $5\#$ is the most and nearly the same as that of UTKFace. In order to balance the data distribution, more CACD images in $5\#$ were augmented by flipping and cropping. Since most CACD images are celebrities with heavy makeup or exaggerated expression, \textit{etc.}, it is difficult to extract the exact age, which results in the lower performance in $5\#$ for our model. 

\begin{table}
\parbox{.45\linewidth}{
\centering
\caption{Age preservation performance of the CAAE and DRAS. \protect\\Bold numbers are the maximums.}
\label{tab:age_cls}
\begin{tabular}{m{2cm}<{\centering}m{2cm}<{\centering}m{2cm}<{\centering}}
\hline
Age Groups & Accuracy of CAAE ($\%$) & Accuracy of DRAS ($\%$)\\
\hline
0$\#$ (0-5) & \textbf{99.94} & 99.93\\
1$\#$ (6-10) & 97.59 & \textbf{99.63}\\
2$\#$ (11-15) & 79.03 & \textbf{92.47}\\
3$\#$ (16-20) & 84.83 & \textbf{93.12}\\
4$\#$ (21-30) & 99.91 & \textbf{100.00}\\
5$\#$ (31-40) & \textbf{97.27} & 93.90\\
6$\#$ (41-50) & 75.64  & \textbf{84.71}\\
7$\#$ (51-60) & \textbf{90.53} & 90.10\\
8$\#$ (61-70) & 96.65 & \textbf{97.02}\\
9$\#$ (70+) & 96.46 &\textbf{99.97}\\
\hline
Average Accuracy($\%$) & 91.78 &\textbf{96.04}\\
\hline
\end{tabular}
}
\hfill
\parbox{.45\linewidth}{
\centering
\caption{Age preservation performance of IPCGAN.}
\label{tab:age_cls_ipcgan}
\begin{tabular}{m{3cm}<{\centering}m{2cm}<{\centering}}
\hline
Age Groups & Accuracy of IPCGAN($\%)$\\
\hline
2$\#$ and 3$\#$ (11-20) & 10.57\\
4$\#$ (21-30) & 25.92\\
5$\#$ (31-40) & 30.89\\
6$\#$ (41-50) & 20.00\\
7$\#$, 8$\#$ and 9$\#$ (50+) & 78.12\\
\hline
Average Accuracy($\%$) & 33.10\\
\hline
\end{tabular}
}
\end{table}

\section{Conclusion}
In this paper we studied a new age synthesis task, namely dual reference age synthesis, and proposed a novel framework. The proposed framework takes two images as inputs, of which one refers to the identity in the target image and the other one refers to the age in the target image. Instead of using a given number as “hard” age information, the DRAS learns the “soft” age information from the age reference image without any age annotations. Compared to the conventional age synthesis methods, the GAN-based DRAS is able to generate higher quality images with more details and closer to the natural effects. Compared to the other GAN-based methods, the DRAS uses an image to describe the age information and doesn’t need pair-wise data for training model. Experimental results on UTKFace and CACD demonstrate the proposed approach showing promising results on this new task.

In this paper, we only consider the Euclidean metric for age preservation loss and identity preservation loss. In fact, a better choice is to minimize the distance between the two probablity distributions of the learnt feature and the reference feature. In our future work, we would like to investigate to use different probablitity divergence metrics as new loss functions. 

\section*{Acknowledgments}
This work was supported by the Jiangsu Overseas Visiting Scholar Program for University Prominent Young and Middle-aged Teachers and Presidents.

\printcredits

\bibliographystyle{elsarticle-num}
\bibliography{cas-refs}

\end{document}